\documentclass[10pt,twocolumn,letterpaper]{article}

\usepackage[pagenumbers]{cvpr} 

\usepackage{setspace}
\usepackage{multirow}
\usepackage{paralist}
\usepackage{multirow}
\usepackage{booktabs}
\usepackage{pifont}
\usepackage{subcaption}
\usepackage{algorithm}
\usepackage{algorithmic}
\usepackage{bbding}
\usepackage{makecell}

\definecolor{iccvblue}{rgb}{0.21,0.49,0.74}
\newcommand{\rf}[1]{{\color{red}{#1}}}
\newcommand{\bd}[1]{{\color{blue}{#1}}} 
\usepackage[pagebackref,breaklinks,colorlinks,allcolors=iccvblue]{hyperref}

\def\paperID{1895} 
\def\confName{CVPR}
\def\confYear{2026}

\title{FoundIR-v2: Optimizing Pre-Training Data Mixtures for \\Image Restoration Foundation Model}

\author{Xiang Chen$^{1}$  \quad Jinshan Pan$^{1}$ \quad Jiangxin Dong$^{1}$ \quad Jian Yang$^{1}$ \quad Jinhui Tang$^{2}$\\
$^{1}$ Nanjing University of Science and Technology \quad $^{2}$ Nanjing Forestry University\\
{\tt \url{https://lowlevelcv.com/}}
}

\begin{document}
\maketitle

\begin{abstract}
Recent studies have witnessed significant advances in image restoration foundation models driven by improvements in the scale and quality of pre-training data.
In this work, we find that the data mixture proportions from different restoration tasks are also a critical factor directly determining the overall performance of all-in-one image restoration models.
To this end, we propose a high-capacity diffusion-based image restoration foundation model, FoundIR-v2, which adopts a data equilibrium scheduling paradigm to dynamically optimize the proportions of mixed training datasets from different tasks.
By leveraging the data mixing law, our method ensures a balanced dataset composition, enabling the model to achieve consistent generalization and comprehensive performance across diverse tasks.
Furthermore, we introduce an effective Mixture-of-Experts (MoE)-driven scheduler into generative pre-training to flexibly allocate task-adaptive diffusion priors for each restoration task, accounting for the distinct degradation forms and levels exhibited by different tasks.
Extensive experiments demonstrate that our method can address over 50 sub-tasks across a broader scope of real-world scenarios and achieves favorable performance against state-of-the-art approaches. 
\end{abstract}

\section{Introduction}
\label{sec:intro}

Recent years have witnessed significant progress in the development of foundation models, which are pre-trained on large-scale data and applied across a wide range of downstream tasks~\cite{foundir,DepthAny}.
Inspired by this popular trend, image restoration foundation models have attracted increasing attention, aiming to simultaneously address extensive restoration tasks within a universal model.

In practice, the efficacy of image restoration foundation models is governed by two critical factors: the composition of training data and the choice of the base model.
The former determines the scope of tasks that the model can learn, whereas the latter dictates its overall learning capability.
Together, these two factors jointly influence the comprehensive performance and robustness of the foundation model.

\begin{figure}[!t]
	\centering
	\includegraphics[width=1.0\columnwidth]{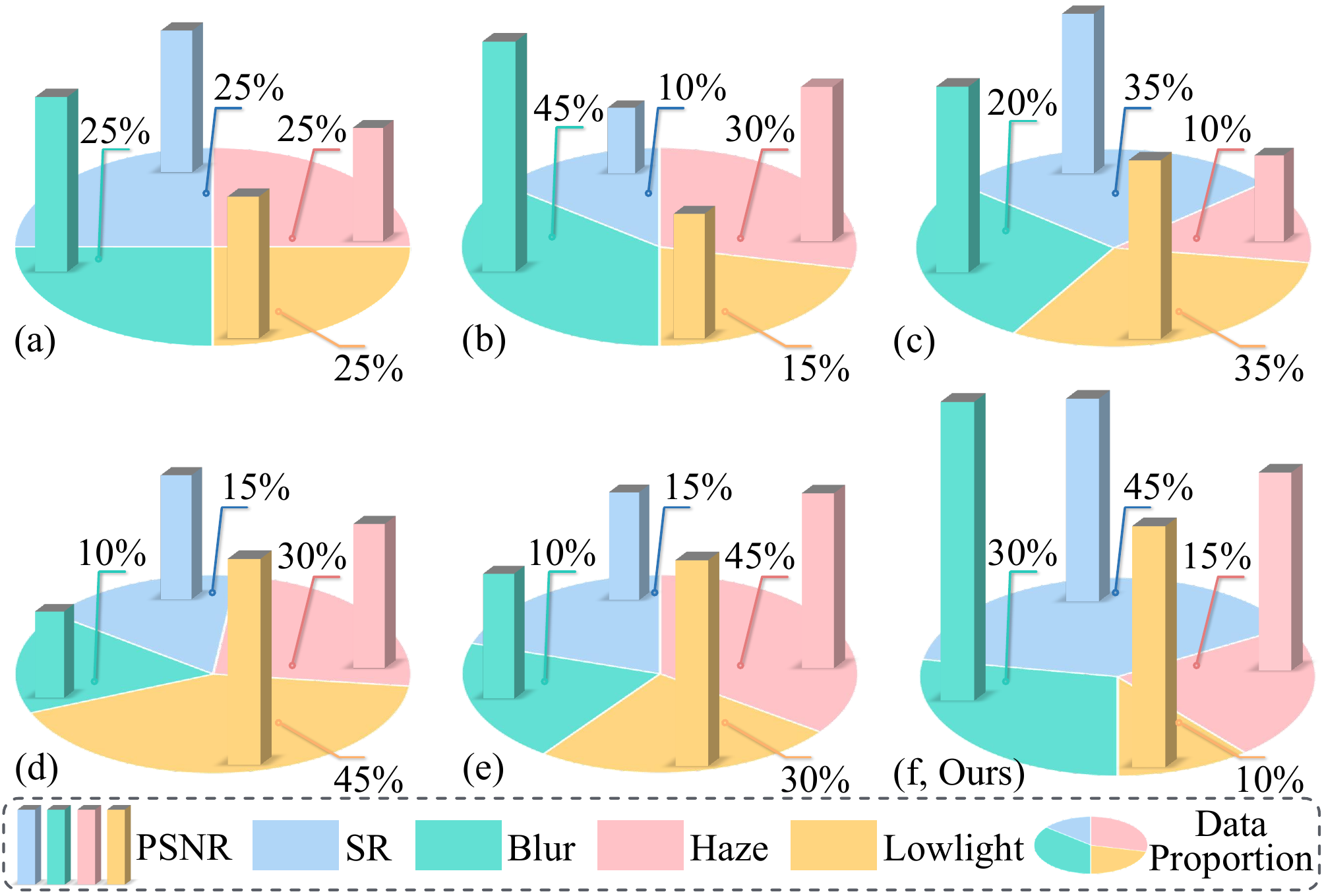}
	\vspace{-6mm}
	\caption{Statistical analysis of the relationship between data mixture proportions and restoration performance for all-in-one foundation model. To facilitate clearer observations, we restrict the experimental analysis to four tasks: deblurring, dehazing, low-light enhancement, and SR. In the figure, the pie chart illustrates the training data distribution across different tasks, while the bar chart reports the corresponding PSNR results on each task’s test set.}
	\label{proportion}
	\vspace{-4mm}
\end{figure}

With respect to \textbf{training data}, existing studies primarily focus on scaling up datasets via data synthesis~\cite{Gendeg} and real-world collection~\cite{foundir} to advance the performance upper bound of universal image restoration.
However, limited attention has been paid to the intricate relationship between the compositional proportions of mixed training datasets (\emph{i.e.}, the relative data volume allocated to different restoration tasks) and the resulting all-in-one restoration capabilities of foundation models.
In fact, the composition of training data from different tasks in a foundation model naturally differs.
In addition, inter-task data interactions during training can exhibit complex relationships that can be facilitative, independent, or even conflicting~\cite{Grids}.
We observe that employing equal or random sampling ratios from different task-specific datasets often leads to considerable performance instability in all-in-one restoration models (see Figure~\ref{proportion}).
In other words, an inappropriate mixture ratio may result in inefficient training or inadequate learning for foundation models.
These insights motivate us to systematically optimize the data composition, with the goal of balancing foundation model capabilities across tasks while leveraging potential synergies among them.

Regarding the \textbf{base framework}, recent diffusion-based approaches have emerged as promising candidates for image restoration foundation model, due to their powerful generative priors that enhance model generalization~\cite{lu2025elucidating}.
However, existing approaches usually either use diffusion models directly~\cite{DiffUIR,IR-SDE,foundir} or fine-tune pretrained ones for all restoration tasks~\cite{SUPIR,lin2024diffbir}, without adequately accounting for the task-dependent impact of diffusion priors under distinct reconstruction objectives.
For example, when processing low-resolution hazy images, current all-in-one restoration methods only perform image dehazing, overlooking the necessity of concurrent super-resolution (SR). This oversight significantly constrains the potential of diffusion priors to achieve high-quality reconstruction.
Thus, this motivates us to adaptively allocate task-specific diffusion priors, improving the joint optimization of various restoration capabilities under dynamic data mixing.

To address these issues, we propose FoundIR-v2, an effective image restoration foundation model capable of handling more than 50 subtasks.
Building upon FoundIR~\cite{foundir}, which investigates the effects of training \textbf{data scaling laws} in image restoration foundation models, this work further uncovers the \textbf{data mixing laws} that govern how mixture proportions influence all-in-one image restoration.
Specifically, we introduce a flexible data equilibrium scheduling paradigm that dynamically adjusts the composition ratios within the training data pool.
This approach enhances the efficacy of data distributions and strengthens the model’s all-in-one restoration ability, thereby catalyzing synergistic interactions across tasks.
In contrast to the static sampling strategies used in existing approaches~\cite{DiffUIR,foundir}, our dynamic data mixing scheme enables the model to obtain comprehensive restoration capabilities more rapidly during the early learning stages, rather than gradually converging at the final stage.
Furthermore, we integrate a mixture-of-experts (MoE)-driven scheduler with Stable Diffusion (SD) to adaptively allocate useful generative priors tailored to the heterogeneous demands of different restoration tasks.
Note that we jointly optimize both data and model scheduling schemes, enabling our method to better align dynamic data mixtures with adaptive model capabilities during generative pre-training.
Extensive experiments show that our FoundIR-v2 outperforms state-of-the-art models in a range of real-world scenarios and downstream applications.

We summarize our main contributions as follows:
\begin{compactitem}
\item We propose FoundIR-v2, an up-to-date image restoration foundation model that dynamically optimizes pre-training data mixtures to better balance the model’s capabilities in all-in-one restoration tasks.

\item We develop an effective MoE-driven diffusion scheduler to dynamically allocate diverse task-adaptive diffusion priors for each restoration tasks during the generative pre-training process.

\item We demonstrate the effectiveness of the data mixing laws, and show that our FoundIR-v2 addresses over 50 sub-tasks across broader real-world scenarios, outperforming state-of-the-art approaches.
\end{compactitem}

\begin{figure*}[t]
    \centering
    \includegraphics[width=\textwidth]{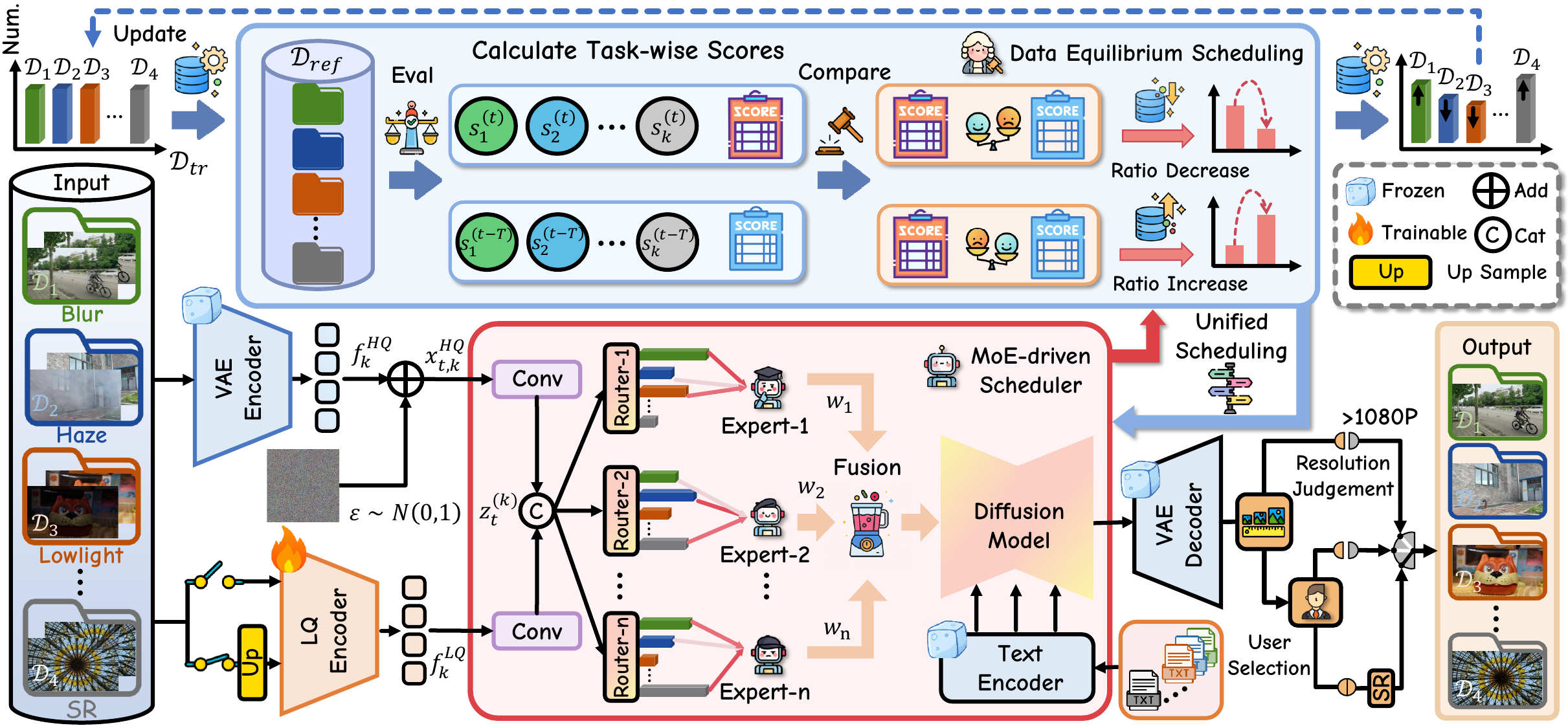}
    \vspace{-6mm}
    \caption{Illustration of the proposed FoundIR-v2. We adopt a dual scheduling strategy for both the training data and the base model, including (i) a data equilibrium scheduling is introduced into the pre-training process to dynamically optimize the data mixture, and (ii) an MoE-driven scheduler is integrated into the model to dynamically allocate task-adaptive diffusion priors. For low-resolution LQ inputs, our proposed FoundIR-v2 allows users to flexibly choose at test time whether to retain the original image resolution or apply SR operation.}
    \label{fig:network}
    \vspace{-4mm}
\end{figure*}

\section{Related Work}
\label{sec:related}

{\flushleft\textbf{Model scaling laws for image restoration}.}
Recent studies on model scaling laws in LLMs~\cite{IDEAL} have demonstrated that increasing model parameters and computational resources leads to more powerful foundation models with improved generalization performance and enhanced reasoning capabilities.
In image restoration, this perspective is being extended to explore effective architectures capable of handling a wide range of real-world degradation tasks within a unified foundation model~\cite{foundir}.
Vision Transformers (ViT) have demonstrated strong potential due to their global receptive fields and scalable design, enabling deeper and wider models for low-level vision. 
For example, Chen \emph{et al}.~\cite{IPT} propose a ViT-based pre-trained large model (IPT) for image processing using 32 NVIDIA Tesla V100 GPUs (32GB).

Recently, diffusion models~\cite{Faithdiff} have emerged as a popular paradigm for image restoration by unleashing generative priors to help model generalization.
SUPIR~\cite{SUPIR} scales up large diffusion models by leveraging a robust large-scale adapter to improve perceptual quality and use text prompts for real-world image restoration.
Later, OmniLV~\cite{pu2025lumina} integrates advanced Diffusion Transformer (DiT)-based generative priors into a multimodal framework, advancing toward general low-level vision~\cite{GenLV}.
More recently, HYPIR~\cite{HYPIR}, trained on 64 NVIDIA A6000 GPUs (48GB), leverages a pre-trained diffusion model for efficient parameter initialization and then applies lightweight adversarial fine-tuning to adapt it specifically for image restoration.
In this paper, we train a robust image restoration foundation model on high-memory NVIDIA H20 GPUs (96 GB), leveraging the great potential of MOE-guided diffusion models.

\vspace{-2mm}

{\flushleft\textbf{Data scaling and mixing laws for image restoration}.}
It is well-known that the training data for foundation models has a significant influence on
their performance.
Li \emph{et al}.~\cite{foundir} find that as the size of real-world training data continues to grow, the performance of all-in-one image restoration models improves substantially.
To this end, FoundIR is proposed by establishing a unified data collection system that enables the acquisition of a million-scale high-quality paired dataset to advance image restoration foundation models~\cite{foundir}.
Furthermore, Rajagopalan \emph{et al}.~\cite{Gendeg} introduce a diffusion-based degradation data synthesis approach that ensures diversity in training data, thereby enhancing the generalization capability of universal image restoration models.
We note that while existing studies focus on increasing the size and quality of training datasets, few works explore how the mixture proportions of different degradation data effect the performance for all-in-one image restoration.
In fact, Data Mixing Law~\cite{IDEAL} introduces a mixing proportion principle derived from extensive experiments in the LLM era, revealing the relationship between data proportions and their corresponding tasks and providing guidance for adjusting data distributions.
In this work, we fill this research gap and propose optimizing data mixtures to better train image restoration foundation models.

\vspace{-2mm}

\section{Proposed Method: FoundIR-v2}
Our goal is to formulate an effective foundation model that can address a broader range of image restoration tasks.
Towards this goal, we propose FoundIR-v2, which maximizes the effectiveness of mixed training datasets and model capacity.
We propose a data equilibrium scheduling scheme and an MoE-driven diffusion scheduler, which are jointly optimized within a unified multi-task framework.
In what follows, we present the details of the proposed approach.

\subsection{Overall framework}
As illustrated in Figure~\ref{fig:network}, our method utilizes a pre-trained Variational Autoencoder (VAE)~\cite{kingma2013auto} encoder to project low-quality (LQ) images into the latent space, producing latent representations denoted as $f^{LQ}$.
Note that the SR task differs from other restoration tasks in that its LQ and HQ images have different image resolutions.
To facilitate unified training for all tasks, for LQ inputs $I_{LQ}$ originating from SR datasets, we apply a random upsampling operation from a set of interpolation techniques, including nearest-neighbor, bilinear, and bicubic interpolation, to align the spatial dimensions of SR inputs with the model’s output resolution.

For the base framework, we adopt the SDXL~\cite{sdxl} backbone as our diffusion model, which is pre-trained on high-quality (HQ) images $I_{HQ}$ to produce the corresponding latent features denoted as $f^{HQ}$.
To better allocate appropriate diffusion priors from SD for each task, we integrate an effective MoE-driven scheduler with the SDXL-based generative pre-training for joint optimization.
In addition, we employ LLAVA~\cite{liu2023visual} to generate image descriptions for all training datasets.
These text embeddings are integrated with the latent features via cross-attention layers to leverage auxiliary text-to-image information for better image restoration~\cite{Faithdiff}.
To ensure adequate task learning and prevent performance imbalance~\cite{foundir}, we incorporate a data equilibrium scheduling scheme into the multi-task learning procedure.
By performing iterative data mixture optimization, we obtain the final outputs using a pre-trained VAE decoder~\cite{kingma2013auto}.

\vspace{-1mm}

\subsection{Data equilibrium scheduling}
{\flushleft\textbf{Problem formulation}.} 
Assume that a base model $\mathcal{M}_0$ with parameters $\theta \in \mathbb{R}^M$ is given. Our final objective is to build upon $\mathcal{M}_0$ to train an effective image restoration foundation model capable of handling diverse sub-tasks.
For the large-scale training data $\mathcal{D}_{tr}$, we partition it into $k$ categories according to task attributes (\emph{e.g.}, image enhancement, super-resolution), where each category is associated with a set of training dataset $\mathcal{D}_{i}$.
Let $N$ denote the total number of training samples, \emph{i.e.}, $N=\left|\mathcal{D}_{t r}\right|=\left|\mathcal{D}_1\right|+\cdots+\left|\mathcal{D}_k\right|$, where $\left|\mathcal{D}_i\right|$ represents the cardinality of the task-specific training dataset $\mathcal{D}_{i}$.
When simply mixing data from multiple tasks, like $\mathcal{D}_{t r}=\mathcal{D}_1 \cup \cdots \cup \mathcal{D}_i$, the foundation model tends to exhibit imbalanced capabilities due to the inappropriate training data distribution.
To facilitate the optimization of training pool, we define the data mixture proportions (weights) $\lambda$ as the sampling probability over the $k$ domains, thereby specifying the distribution of training data:
\vspace{-8pt} 
\begin{equation}
P_\lambda=\sum_{i=1}^k \lambda_i \operatorname{unif}\left(\mathcal{D}_{i}\right),
\end{equation}

\vspace{-8pt} 
\noindent where $\operatorname{unif}(\mathcal{D})=\frac{1}{|\mathcal{D}|} \sum_{x \in \mathcal{D}} \delta_x$ is the uniform distribution over the image samples in $\mathcal{D}$, and $\delta_x$ denotes the Dirac delta function centered at sample $x$.
Given a fixed model size, the training seeks to obtain the optimal model parameters $\theta^*$ by minimizing the L1-normalized function:
\begin{equation}
\theta_\lambda^*=\underset{{\theta}}{\arg \min } \parallel I_{HQ}- \mathcal{M}_0(I_{LQ})\parallel_1,
\end{equation}
where $\parallel \cdot \parallel_1$ denotes the L1-normalized reconstruction loss.

\vspace{-2mm}

{\flushleft\textbf{Optimization of data mixture}.} 
We note that existing methods~\cite{DiffUIR,foundir,DA-CLIP} typically adopt statically fixed data mixing ratios for model training.
To maximize the effectiveness of mixed training dataset, we dynamically optimize the data mixture by redefining the ratios at scheduled intervals during the training iterations.
Specifically, we first construct an initial training batch by uniformly sampling from the subsets $\left\{\mathcal{D}_1, \ldots, \mathcal{D}_i\right\}$  of $\mathcal{D}_{tr}$.
After every interval of $T$ iterations, the performance of the current model is evaluated on a small independent reference dataset $\mathcal{D}_{ref}$ to validate its multi-task restoration performance.
Here, $\mathcal{D}_{ref}$ is randomly sampled from $\mathcal{D}_{tr}$ across $k$ categories, with an equal number of samples per task, and it is disjoint from $\mathcal{D}_{tr}$.
Based on this evaluation, we obtain task-wise performance scores ${s_1^{(t)},\ldots,s_k^{(t)}}$ at iteration $t$ on $\mathcal{D}_{ref}$.
To track multi-task restoration performance trends, we compare them with those at the previous checkpoint ${s_1^{(t-T)},\ldots,s_k^{(t-T)}}$.
If the performance of task $j$ degrades, i.e., $s_j^{(t)} < s_j^{(t-T)}$, we increase its data sampling probability in the next training interval to allocate more supervision from $\mathcal{D}_j$.
Conversely, if a task maintains or improves performance, its weight can be reduced slightly to balance the overall data distribution.

Formally, we define the rule for data optimization as
\begin{equation}
\lambda_j^{(t+1)} = \frac{\lambda_j^{(t)}\exp\left(-\alpha \Delta s_j^{(t)}\right)}{\sum_{i=1}^k \lambda_i^{(t)} \exp\left(-\alpha \Delta s_i^{(t)}\right)},
\end{equation}
where $\Delta s_j^{(t)} = s_j^{(t)} - s_j^{(t-T)}$ denotes the performance change of task $j$, and $\alpha > 0$ is a scaling factor controlling the sensitivity of composition proportion adjustment.
This softmax-based normalization ensures $\sum_{j=1}^k \lambda_j^{(t+1)} = 1$, while reweighting the sampling probabilities across tasks.

The pseudocode for data equilibrium scheduling is presented in Algorithm~\ref{alg:DES}. 
Through this dynamic optimization scheme, our method allocates more data to underperforming tasks while preventing overfitting on already well-learned tasks, thereby mitigating task imbalance and improving the overall generalization of the foundation model.

\setlength{\textfloatsep}{8pt}

\begin{algorithm}[t]
\caption{Data Equilibrium Scheduling}
\label{alg:DES}
\begin{algorithmic}[1]
\REQUIRE Initial model $\mathcal{M}_0$, task categories $k$, training dataset ${D}_{tr} = \{\mathcal{D}_{tr}^{1},...,\mathcal{D}_{tr}^{k}\}$, reference dataset $\mathcal{D}_{ref}=\{\mathcal{D}_{ref}^{1},...,\mathcal{D}_{ref}^{k}\}$, evaluation interval $T$, maximum iterations $I$, sample batches $\{\mathcal{D}_{1},...,\mathcal{D}_{k}\}^{t} \in {D}_{tr}$.

\FOR{$t = 1$ to $I$}
\STATE Train model $\mathcal{M}_t$ on sample batches $\{\mathcal{D}_{1},...,\mathcal{D}_{k}\}^{t}$;
    \IF{$t \bmod T = 0$}
        \STATE Evaluate $\mathcal{M}_t$ on $\mathcal{D}_{ref}=\{\mathcal{D}_{ref}^{1},...,\mathcal{D}_{ref}^{k}\}$ to obtain current scores $ \{s_{1}^{t},...,s_{k}^{t}\}$ for all tasks $k$; 

        \FOR{$j = 1$ to $k$}
            \IF{$s_{j}^{t} < s_{j}^{(t-T)}$} 
                \STATE Increase data sampling proportion $\lambda_j^{(t+T)}$;
            \ELSE
                \STATE Decrease data sampling proportion $\lambda_j^{(t+T)}$;
            \ENDIF
        \ENDFOR
    \ENDIF
    \STATE Update sample batches $\{\mathcal{D}_{1},...,\mathcal{D}_{k}\}^{(t+T)} \in {D}_{tr}$;
\ENDFOR
\STATE \textbf{return} Final model $\mathcal{M}_{I}$.
\end{algorithmic}
\end{algorithm}

\vspace{-1mm}

\subsection{MoE-driven diffusion scheduler}
Since diverse image restoration tasks involve heterogeneous reconstruction demands in multi-task learning, we integrate an MoE-driven scheduler into the generative pre-training to allocate task-adaptive diffusion priors.
Specifically, given the LQ feature $f^{L Q}$ and the noisy latent $x_{t}^{H Q}$ from $k$ tasks generated by the diffusion model at timestep $t$, we first obtain a fused representation $\mathbf{z}_t^{(k)}=\phi\left(f_k^{L Q}, x_{t, k}^{H Q}\right)$, where $\phi(\cdot)$ denotes the concatenation operation.
Then, this representation is fed into an MoE-driven scheduler consisting of $n$ shared experts to capture diverse degradation patterns.
By the router's dynamic selection, the associated weights $w_i^{(k)}$ generated by different experts are defined as follows:
\begin{equation}
w_i^{(k)}=\frac{\exp \left(\mathbf{g}_i^{(k) \top} \mathbf{z}_t^{(k)}\right)}{\sum_{j=1}^n \exp \left(\mathbf{g}_j^{(k) \top} \mathbf{z}_t^{(k)}\right)},
\end{equation}
where $\mathbf{g}_i^{(k)}$ is the learnable gating parameter for each expert. This formulation converts the scores of all experts into non-negative weights that sum to 1, enabling a weighted combination of the outputs from all experts.

Here, each expert $E_i(\cdot)$ denotes a type of attention mechanism (\emph{e.g.}, spatial~\cite{vaswani2017attention}, channel~\cite{Restormer}, or sparse~\cite{DRSformer}), designed to activate different cues based on task requirements.
By executing model scheduling, our method can effectively exploit useful information from the degraded inputs across different tasks, thereby facilitating the subsequent diffusion process.
Finally, we obtain the scheduled feature representation $\mathcal{F}^{(k)}$, which is defined as follows:
\vspace{-6pt}
\begin{equation}
\mathcal{F}^{(k)}\left(z_t\right)=\sum_{i=1}^n w_i^{(k)} E_i\left(z_t^{(k)}\right).
\end{equation}
\vspace{-3mm}

Inspired by~\cite{Faithdiff}, we first pre-train only the MoE-driven scheduler, and then jointly fine-tune the scheduler and the diffusion model, in order to better enhance the consistency of multi-task feature utilization for generative pre-training.

\vspace{-2mm}

\section{Experiments}

\subsection{Experimental settings}

\begin{figure}[!t]
\centering
\includegraphics[width=1.0\columnwidth]{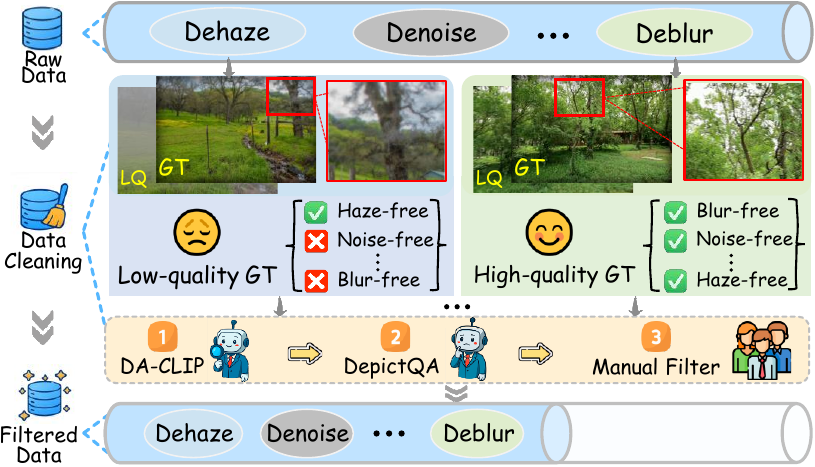}
\vspace{-6mm}
\caption{Visual example of high-quality GT data filtering.}
\label{fig:dataset}
\end{figure}

{\flushleft\textbf{Training dataset}.}  
We combine publicly available datasets from a diverse set of image restoration tasks to form our training dataset~\cite{foundir,weatherbench,4KRD,MC-Blur,LSDIR,DIV2K,Flickr2K,DIV8K,FFHQ,Uav-rain1k,chen2025towards}.
We note that existing studies only focus on the quality of LQ images in the training dataset (\emph{i.e.}, data diversity of different degradation types), but ignore the impact of GT image quality on all-in-one restoration performance.
In other words, existing approaches simply combine the GT images from individual restoration tasks, while overlooking the common characteristics of clear images across all tasks.
For example, Figure~\ref{fig:dataset} shows a visual example from the dehazing task in the existing training set. Although its GT image is haze-free, it contains other visible degradation types, such as blur and noise.
This would lead to negative conflicts in learning objectives during the data mixing training, potentially interfering with the model’s generalization performance.

To alleviate this problem, we employ recent multi-modal image quality assessment (IQA) models~\cite{DA-CLIP,DepictQA} to perform degradation identification and quality evaluation on the GT images within the training dataset. 
By leveraging the IQA models, we further filter out low-quality GT samples, ensuring that only high-quality data are retained for training.
This data cleaning process can mitigate potential data contamination effects during pre-training data mixture.

\vspace{-2mm}

{\flushleft\textbf{Test datasets}.}  
To comprehensively evaluate the generalization ability of our model in all-in-one image restoration, we test the model on a wide range of benchmark datasets across diverse tasks, \emph{i.e.}, 4KRD~\cite{4KRD}, LSD~\cite{MC-Blur}, PolyU~\cite{PolyU}, HQ-NightRain~\cite{CSTNet}, UAV-Rain1k~\cite{Uav-rain1k}, WeatherBench~\cite{weatherbench}, Dense-HAZE~\cite{Dense-haze}, NH-HAZE~\cite{NH-HAZE}, RS-Cloud~\cite{ning2025cloud}, UHD-LL~\cite{UHD-LL}, FoundIR-TestData~\cite{foundir}, RealPhoto60~\cite{SUPIR}, and RealDeg~\cite{Faithdiff}.
More experimental results on other restoration tasks are included in the supplementary material.

\vspace{-2mm}

{\flushleft\textbf{Evaluation metrics}.}  
We evaluate the fidelity and quality of restored images using PSNR, SSIM, and a set of perceptual metrics (\emph{i.e.}, LPIPS~\cite{LPIPS}, MUSIQ~\cite{MUSIQ}, CLIPIQA+~\cite{CLIPIQA}, PIQE~\cite{PIQE}, MANIQA~\cite{MANIQA} and PaQ-2-PiQ~\cite{PaQ-2-PiQ}).

\vspace{-2mm}

{\flushleft\textbf{Implementation details}.}  
We train our proposed FoundIR-v2 on two NVIDIA H20 GPUs (96 GB) using the AdamW optimizer~\cite{loshchilov2017decoupled} with default parameters.
During training, images are randomly cropped into $512 \times 512$ patches, with a batch size of 32.
We follow the unified feature optimization strategy in~\cite{Faithdiff}. The initial learning rate of the VAE encoder is set to $5 \times 10^{-6}$, while that of the other components is set to $5 \times 10^{-5}$.
The learning rate is subsequently scheduled by the cosine annealing scheme~\cite{loshchilov2016sgdr}.
We train the network for a total of 150,000 iterations.
During the validation, MUSIQ is employed to calculate task-wise scores for the SR task, whereas PSNR is used for other tasks.
For $\mathcal{D}_{ref}$, the number of validation samples for each task category is 10.
During the model inference, we adopt the Euler scheduler~\cite{karras2022elucidating} with 20 sampling timesteps for all tasks, while the classifier-free guidance (CFG)~\cite{ho2022classifier} scale is fixed at 5.
For the SR task, we employ AdaIN as the color-fix strategy, while no color-fix is applied to the other tasks.
The code will be available.

\begin{table*}[h!]
\centering
\caption{Quantitative comparisons with state-of-the-art general restoration methods, all-in-one restoration models, image restoration agent, and real-world SR approaches on benchmark datasets. \rf{Red} and \bd{Blue} indicate the best and the second-best performance.}
\vspace{-2mm}
\resizebox{\textwidth}{!}{%
\begin{tabular}{l|ccccc||ccccc||ccccc}
\toprule

\multirow{2}{*}{Methods} & \multicolumn{5}{c||}{\textbf{4KRD (Motion Deblurring)}~\cite{4KRD}}     & \multicolumn{5}{c||}{\textbf{LSD (Defocus Deblurring)}~\cite{MC-Blur}}     & \multicolumn{5}{c}{\textbf{PolyU (Denoising)}~\cite{PolyU}}  \\
                         & PSNR~$\uparrow$    & SSIM~$\uparrow$    & LPIPS~$\downarrow$   & MUSIQ~$\uparrow$   & CLIPIQA+~$\uparrow$ & PSNR~$\uparrow$    & SSIM~$\uparrow$    & LPIPS~$\downarrow$   & MUSIQ~$\uparrow$   & CLIPIQA+~$\uparrow$ & MUSIQ~$\uparrow$    & CLIPIQA+~$\uparrow$    & PIQE~$\downarrow$   & MANIQA~$\uparrow$   & PaQ-2-PiQ~$\uparrow$   \\ \hline
PromptIR~\cite{PromptIR}  & 24.74 & 0.7678 & 0.3386 & 21.02 & 0.3232 & 19.63 & 0.7335 & 0.5582 & 15.04 & 0.2225 & \bd{51.80} & 0.3873 & 60.3508 & 0.5182 & 68.7471   \\
TransWeather~\cite{Transweather}     & 23.36 & 0.7231 & 0.4384 & 17.78 & 0.2563 & 19.87 & 0.7308 & 0.5563 & 14.85 & 0.2468 & 32.99 & 0.3111 & 80.8918 & 0.4030 & 59.8569   \\
DA-CLIP~\cite{DA-CLIP}   & 23.84 & 0.7347 & 0.4408 & 18.13 & 0.3103 & 19.65 & 0.7313 & 0.5179 & 15.73 & 0.2178 & 42.13 & 0.3547 & 74.2320 & 0.4774 & 65.3832   \\
DiffUIR~\cite{DiffUIR}   & 24.69 & 0.7596 & 0.3562 & 20.22 & 0.3225 & 19.71 & \bd{0.7357} & 0.5818 & 16.05 & 0.2071 & 45.53 & 0.3561 & 73.2232 & 0.4611 & 66.6793  \\
AutoDIR~\cite{AutoDIR}   & 24.21 & 0.7288 & 0.4396 & 18.42 & 0.2917 & \bd{20.01} & \rf{0.7431} & \bd{0.5163} & 18.82 & 0.2389 & 37.67 & 0.3538 & 80.3762 & 0.4496 & 63.2148   \\
InstructIR~\cite{InstructIR} & 26.28 & \bd{0.8181} & \bd{0.2307} & \bd{27.84} & \bd{0.3844} & 19.28 & 0.7227 & 0.5220 & 15.82 & 0.2121 & \rf{52.14} & \rf{0.4445} & \bd{40.8439} & 0.5254 & \bd{70.7628}   \\
X-Restormer~\cite{X-Restormer}  & 24.57 & 0.7604 & 0.3823 & 19.97 & 0.3188 & 19.46 & 0.7336 & 0.5854 & 15.84 & 0.2288 & 32.99 & 0.3111 & 80.8918 & 0.4030 & 59.8569   \\
AgenticIR~\cite{AgenticIR} & 24.41 & 0.7569 & 0.4021 & 18.73 & 0.3196 & 18.30 & 0.7146 & 0.5337 & \bd{22.63} & \bd{0.2608} & 35.98 & 0.3186 & 79.2196 & 0.4580 & 62.2334   \\
FoundIR~\cite{foundir}   & \bd{26.59} & \rf{0.8188} & 0.2351 & 27.35 & 0.3676 & 19.18 & 0.7306 & 0.5701 & 15.92 & 0.2007 & 49.92 & 0.4244 & 42.0997 & \bd{0.5293} & \rf{70.9032}   \\
Ours                     & \rf{26.64} & 0.7997 & \rf{0.1451} & \rf{34.93} & \rf{0.4796} & \rf{20.78} & 0.6708 & \rf{0.3658} & \rf{50.74} & \rf{0.3615}  & 48.68 & \bd{0.4276} & \rf{31.7934} & \rf{0.5497} & 70.2896   \\ \hline

\multirow{2}{*}{Methods} & \multicolumn{5}{c||}{\textbf{Dense-HAZE (Dehazing)}~\cite{Dense-haze}}     & \multicolumn{5}{c||}{\textbf{NH-HAZE (Non-Homogeneous Dehazing)}~\cite{NH-HAZE}}     & \multicolumn{5}{c}{\textbf{RS-Cloud (Declouding)}~\cite{ning2025cloud}}  \\
                         & PSNR~$\uparrow$    & SSIM~$\uparrow$    & LPIPS~$\downarrow$   & MUSIQ~$\uparrow$   & CLIPIQA+~$\uparrow$ & PSNR~$\uparrow$    & SSIM~$\uparrow$    & LPIPS~$\downarrow$   & MUSIQ~$\uparrow$   & CLIPIQA+~$\uparrow$ & PSNR~$\uparrow$    & SSIM~$\uparrow$    & LPIPS~$\downarrow$   & MUSIQ~$\uparrow$   & CLIPIQA+~$\uparrow$   \\ \hline
PromptIR~\cite{PromptIR}  & 9.57 &  0.4333 & 0.7889 & 26.16 & 0.2406 & 11.38 & 0.4343 & 0.6077 & 50.40 & 0.3288 & 11.35 & 0.7716 & 0.2713 & 41.83 & \bd{0.4739}   \\
TransWeather~\cite{Transweather}     & 10.51 & 0.4523 & 0.7900 & 27.85 & 0.2060 & 11.58 & 0.4110 & 0.6921 & 40.22 & 0.2543 & 12.93 & 0.7081 & 0.3808 & 30.02 & 0.3390   \\
DA-CLIP~\cite{DA-CLIP}   & 10.94 & 0.4590 & 0.7604 & 30.07 & 0.2451 & 12.35 & 0.4662 & 0.5903 & 49.73 & 0.3364 & 16.43 & 0.8028 & 0.2383 & 35.73 & \rf{0.4839}   \\
DiffUIR~\cite{DiffUIR}   & 9.59 & 0.4326 & 0.7994 & 26.42 & \bd{0.2680} & 11.39 & 0.4220 & 0.6508 & 47.77 & 0.3068 & 13.82 & 0.8140 & 0.2602 & 38.44 & 0.4210   \\
AutoDIR~\cite{AutoDIR}   & \bd{12.33} & \bd{0.4862} & \bd{0.7364} & 31.97 & 0.2167 & \bd{12.71} & \bd{0.4774} & 0.6119 & 46.29 & 0.2991 & \bd{18.39} & 0.8093 & 0.2519 & 31.44 & 0.4390   \\
InstructIR~\cite{InstructIR} & 11.03 & 0.4649 & 0.7456 & \bd{32.96} & 0.1615 & 12.24 & \rf{0.4984} & \bd{0.5303} & \bd{54.80} & 0.2455 & 14.46 & \rf{0.8647} & \bd{0.1624} & \rf{47.00} & 0.3538 \\
X-Restormer~\cite{X-Restormer}  & 9.57 & 0.4295 & 0.8049 & 24.09 & 0.2486  & 11.36 & 0.4131 & 0.6653 & 44.28 & 0.2967 & 11.48 & 0.7506 & 0.3252 & 35.99 & 0.3641  \\
AgenticIR~\cite{AgenticIR} & 10.11 & 0.3884 & 0.7960 & 27.73 & 0.2595 & 12.20 & 0.4495 & 0.6745 & 37.47 & \bd{0.3664} & 17.80 & 0.7992 & 0.2572 & 34.90 & 0.4297   \\
FoundIR~\cite{foundir}   & 9.29 & 0.4307 & 0.7951 & 27.80 & 0.2522 & 11.43 & 0.4491 & 0.5817 & 54.53 & 0.2978 & 11.71 & 0.7527 & 0.3219 & 35.40 & 0.4295   \\
Ours                     & \rf{15.29} & \rf{0.5063} & \rf{0.4975} & \rf{57.48} & \rf{0.3672} & \rf{17.00} & 0.4616 & \rf{0.3338} & \rf{62.62} & \rf{0.4065}  & \rf{22.06} & \bd{0.8278} & \rf{0.1251} & \bd{44.37} & 0.4411   \\ \hline

\multirow{2}{*}{Methods} & \multicolumn{5}{c||}{\textbf{HQ-NightRain (Deraining)}~\cite{CSTNet}}     & \multicolumn{5}{c||}{\textbf{UAV-Rain1k (Raindrop Removal)}~\cite{Uav-rain1k}}     & \multicolumn{5}{c}{\textbf{WeatherBench (Desnowing)}~\cite{weatherbench}}  \\
                         & PSNR~$\uparrow$    & SSIM~$\uparrow$    & LPIPS~$\downarrow$   & MUSIQ~$\uparrow$   & CLIPIQA+~$\uparrow$ & PSNR~$\uparrow$    & SSIM~$\uparrow$    & LPIPS~$\downarrow$   & MUSIQ~$\uparrow$   & CLIPIQA+~$\uparrow$ & PSNR~$\uparrow$    & SSIM~$\uparrow$    & LPIPS~$\downarrow$   & MUSIQ~$\uparrow$   & CLIPIQA+~$\uparrow$   \\ \hline
PromptIR~\cite{PromptIR}  & 10.43 & 0.4485 & 0.5684 & 38.70 & 0.3727 & 15.16 & \rf{0.6605} & \bd{0.4025} & 64.44 & 0.5712 & 21.54 & 0.7767 & 0.2421 & 45.20 & 0.3622   \\
TransWeather~\cite{Transweather}     & 12.88 & 0.4774 & 0.6642 & 28.58 & 0.2855 & 14.85 & 0.5381 & 0.5984 & 51.68 & 0.4145 & 20.94 & 0.7579 & 0.2531 & 44.25 & 0.3306   \\
DA-CLIP~\cite{DA-CLIP}   & 14.78 & 0.5398 & 0.5717 & 32.11 & 0.3500 & 15.38 & 0.6035 & 0.4414 & 62.41 & 0.5867 & 21.59 & 0.7731 & \rf{0.2357} & 45.22 & 0.3624   \\
DiffUIR~\cite{DiffUIR}   & 11.58 & 0.4827 & 0.5687 & 34.57 & 0.3277 & 15.20 & 0.6389 & 0.4362 & 63.93 & 0.5448 & \bd{21.68} & \bd{0.7795} & \bd{0.2399} & 45.39 & 0.3648   \\
AutoDIR~\cite{AutoDIR}   & 11.67 & 0.4697 & 0.6114 & 31.65 & \rf{0.3766} & \bd{15.41} & 0.5834 & 0.5079 & 59.65 & 0.5552 & - & - & - & - & -   \\
InstructIR~\cite{InstructIR} & 10.92 & 0.3972 & 0.4802 & \rf{46.46} & 0.3475 & 13.75 & 0.3240 & 0.4344 & \bd{64.62} & \rf{0.6213} & - & - & - & - & -   \\
X-Restormer~\cite{X-Restormer}  & 10.33 & 0.4376 & 0.6093 & 36.12 & 0.3420 & 15.16 & 0.6397 & 0.4339 & 61.18 & 0.5468 & 21.57 & \rf{0.7813} & 0.2504 & 46.04 & 0.3520   \\
AgenticIR~\cite{AgenticIR} & \bd{16.29} & \bd{0.5681} & 0.5797 & 32.16 & \bd{0.3762} & 14.26 & 0.5300 & 0.5692 & 54.13 & 0.5457 & 20.35 & 0.7304 & 0.2831 & \bd{50.87} & \bd{0.3964}   \\
FoundIR~\cite{foundir}   & 11.57 & 0.5220 & \bd{0.4708} & 41.68 & 0.3570 & 15.11 & \bd{0.6411} & 0.4296 & 64.18 & 0.5550 & 21.57 & 0.7794 & 0.2407 & 45.72 & 0.3554   \\
Ours                     & \rf{18.57} & \rf{0.6083} & \rf{0.2419} & \bd{44.29} & 0.3312 & \rf{17.90} & 0.5147 & \rf{0.2373} & \rf{67.69} & \bd{0.5884}  & \rf{23.15} & 0.7062 & 0.2538 & \rf{60.15} & \rf{0.4496}   \\ \hline

\multirow{2}{*}{Methods} & \multicolumn{5}{c||}{\textbf{UHD-LL (UHD Enhancement)}~\cite{UHD-LL}}     & \multicolumn{5}{c||}{\textbf{FoundIR-L (Low-light Enhancement)}~\cite{foundir}}     & \multicolumn{5}{c}{\textbf{FoundIR-L+N (Joint Denoising and Enhancement)}~\cite{foundir}}  \\
                         & PSNR~$\uparrow$    & SSIM~$\uparrow$    & LPIPS~$\downarrow$   & MASIQ~$\uparrow$   & CLIPIQA+~$\uparrow$ & PSNR~$\uparrow$    & SSIM~$\uparrow$    & LPIPS~$\downarrow$   & MUSIQ~$\uparrow$   & CLIPIQA+~$\uparrow$ & PSNR~$\uparrow$    & SSIM~$\uparrow$    & LPIPS~$\downarrow$   & MUSIQ~$\uparrow$   & CLIPIQA+~$\uparrow$   \\ \hline
PromptIR~\cite{PromptIR}  & 11.76 & 0.6170 & 0.4554 & 28.24 & 0.2075 & 16.33 & 0.6371 & 0.4871 & 27.62 & 0.2634 & 10.83 & 0.4558 & 0.6791 & 28.13 & 0.2543   \\
TransWeather~\cite{Transweather}     & 12.30 & 0.6557 & 0.4978 & 23.01 & 0.2004 & 14.95 & 0.6353 & 0.5747 & 22.31 & 0.2021 & 11.56 & 0.5256 & 0.6969 & 23.57 & 0.2081  \\
DA-CLIP~\cite{DA-CLIP}   & 18.51 & 0.8120 & 0.4130 & 23.45 & \bd{0.2807} & 17.34 & 0.7508 & 0.5067 & 25.75 & 0.2849 & 15.70 & 0.6457 & 0.6563 & 24.27 & 0.2978   \\
DiffUIR~\cite{DiffUIR}   & 11.27 & 0.5975 & 0.4485 & 26.07 & 0.2255 & 14.02 & 0.7172 & 0.3070 & 47.66 & \bd{0.3838} & 13.88 & 0.6394 & 0.4514 & 45.58 & 0.3858   \\
AutoDIR~\cite{AutoDIR}   & \rf{22.52} & \rf{0.8572} & 0.3919 & 21.59 & 0.2453 & \rf{21.91} & 0.8385 & 0.3563 & 31.55 & 0.3364 & \bd{17.50} & 0.7007 & 0.5023 & 30.18 & 0.2821   \\   
InstructIR~\cite{InstructIR} & 20.03 & 0.7356 & 0.4192 & \bd{34.21} & 0.2324 & \bd{20.04} & \rf{0.8665} & \bd{0.2660} & 47.83 & 0.3109 & 16.73 & 0.4719 & 0.6228 & 33.05 & 0.2197   \\
X-Restormer~\cite{X-Restormer}  & 11.56 & 0.6113 & 0.4788 & 26.13 & 0.2116 & 16.02 & 0.6432 & 0.5034 & 27.48 & 0.2506 & 9.75 & 0.3925 & 0.7185 & 27.62 & 0.2396   \\
AgenticIR~\cite{AgenticIR} & 12.82 & 0.6649 & 0.4961 & 24.07 & 0.2208 & 8.51 & 0.3816 & 0.5595 & 28.09 & 0.2754 & 10.73 & 0.5103 & 0.6917 & 27.40 & 0.2696    \\
FoundIR~\cite{foundir}   & 10.61 & 0.5775 & \bd{0.3692} & 32.91 & 0.2394 & 18.98 & 0.8372 & 0.2739 & \bd{49.00} & 0.3567 & 16.12 & \bd{0.7272} & \bd{0.4003} & \bd{50.86} & \bd{0.4552}   \\
Ours                     & \bd{20.08} & \bd{0.8484} & \rf{0.2334} & \rf{45.24} & \rf{0.3575} & 19.72 & \bd{0.8494} & \rf{0.1714} & \rf{56.43} & \rf{0.4009}  & \rf{19.72} & \rf{0.7425} & \rf{0.2652} & \rf{61.33} & \rf{0.4762}   \\ \hline
\multirow{2}{*}{Methods} & \multicolumn{5}{c||}{\textbf{RealPhoto60 (Super-Resolution)}~\cite{SUPIR}}     & \multicolumn{5}{c||}{\textbf{RealDeg (Old Photo Restoration)}~\cite{Faithdiff}}     & \multicolumn{5}{c}{\textbf{RealDeg (Face Restoration)}~\cite{Faithdiff}}          \\
                         & MUSIQ~$\uparrow$    & CLIPIQA+~$\uparrow$    & PIQE~$\downarrow$   & MANIQA~$\uparrow$   & PaQ-2-PiQ~$\uparrow$ & MUSIQ~$\uparrow$    & CLIPIQA+~$\uparrow$    & PIQE~$\downarrow$   & MANIQA~$\uparrow$   & PaQ-2-PiQ~$\uparrow$ & MUSIQ~$\uparrow$    & CLIPIQA+~$\uparrow$    & PIQE~$\downarrow$   & MANIQA~$\uparrow$   & PaQ-2-PiQ~$\uparrow$   \\ \hline
Real-ESRGAN~\cite{Real-esrgan}         & 59.29 & 0.4398 & 25.0258 & 0.5046 & 69.0438 & 54.51 & 0.3700 & 22.5213 & 0.4972 & 67.8305 & 51.66 & 0.2947 & 26.2679 & 0.4541 & 69.5341   \\
BSRGAN~\cite{BSRGAN}  & 45.46 & 0.3397 & - & 0.3759 & 63.3800 & 40.59 & 0.2956 & 35.5768 & 0.4008 & 63.9088 & 53.19 & 0.3429 & 26.8261 & 0.4486 & 70.0197   \\
StabeSR~\cite{StabeSR}  & 52.65 & 0.3916 & 27.1920 & 0.4361 & 64.3983 & - & - & - & - & - & - & - & - & - & -   \\
PASD~\cite{PASD}                 & 63.53 & 0.4787 & 29.2662 & 0.5194 & 69.9744 & 34.59 & 0.2386 & 45.7651 & 0.3990 & 61.4419 & 36.97 & 0.2479 & 62.6459 & 0.3913 & 61.5455    \\
SeeSR~\cite{Seesr}                 & 71.74 & \bd{0.5956} & 26.7787 & 0.6077 & 73.8402 & \bd{59.52} & 0.3596 & 29.1688 & 0.5115 & 69.9719 & 56.58 & 0.3489 & 28.5668 & 0.4509 & 70.5475  \\
DreamClear~\cite{DreamClear}             & 70.46 & 0.5273 & 26.0232 & 0.6080 & \rf{75.8422} & 52.69 & \bd{0.4024} & 24.6792 & 0.5259 & 68.6876 & 54.23 & 0.3385 & 21.5950 & 0.4849 & 69.3709  \\
OSEDiff~\cite{OSEDiff} & 70.46 & 0.5725 & 28.6904 & 0.5892 & 73.4461 & - & - & - & - & - & - & - & - & - & -  \\
SUPIR~\cite{SUPIR}      & 70.26 & 0.5527 & \bd{23.2780} & 0.6064 & 74.4188 & 53.30 & 0.3152 & \rf{18.8356} & 0.5313 & 69.9300 & 53.17 & 0.3179 & \bd{20.6981} & 0.4823 & 69.6014  \\
FaithDiff~\cite{Faithdiff}  & \rf{72.74} & 0.5933 & 25.0027 & \bd{0.6430} & \bd{75.2664} & 53.32 & 0.3420 & \bd{19.1189} & \bd{0.5452} & \bd{70.1377} & \bd{59.95} & \bd{0.3497} & \rf{20.3812} & \bd{0.5216} & \rf{71.8938}  \\
Ours                     & \bd{72.36} & \rf{0.5977} & \rf{23.2547} & \rf{0.6441} & 75.0014 & \rf{62.72} & \rf{0.4730} & 33.3370 & \rf{0.5928} & \rf{71.9529} & \rf{64.03} & \rf{0.4460} & 21.7626 & \rf{0.5481} & \bd{71.7996} \\ \bottomrule
\end{tabular}
}
\label{tab:comparison}
\vspace{-5mm}
\end{table*}

\subsection{Comparisons with the state of the art}

\begin{figure*}[t]
	\footnotesize
	\begin{center}
		\begin{tabular}{c c c c c c c c}
			 \multicolumn{3}{c}{\multirow{5}*[48pt]{
            \hspace{-2.5mm} \includegraphics[width=0.42\linewidth,height=0.245\linewidth]{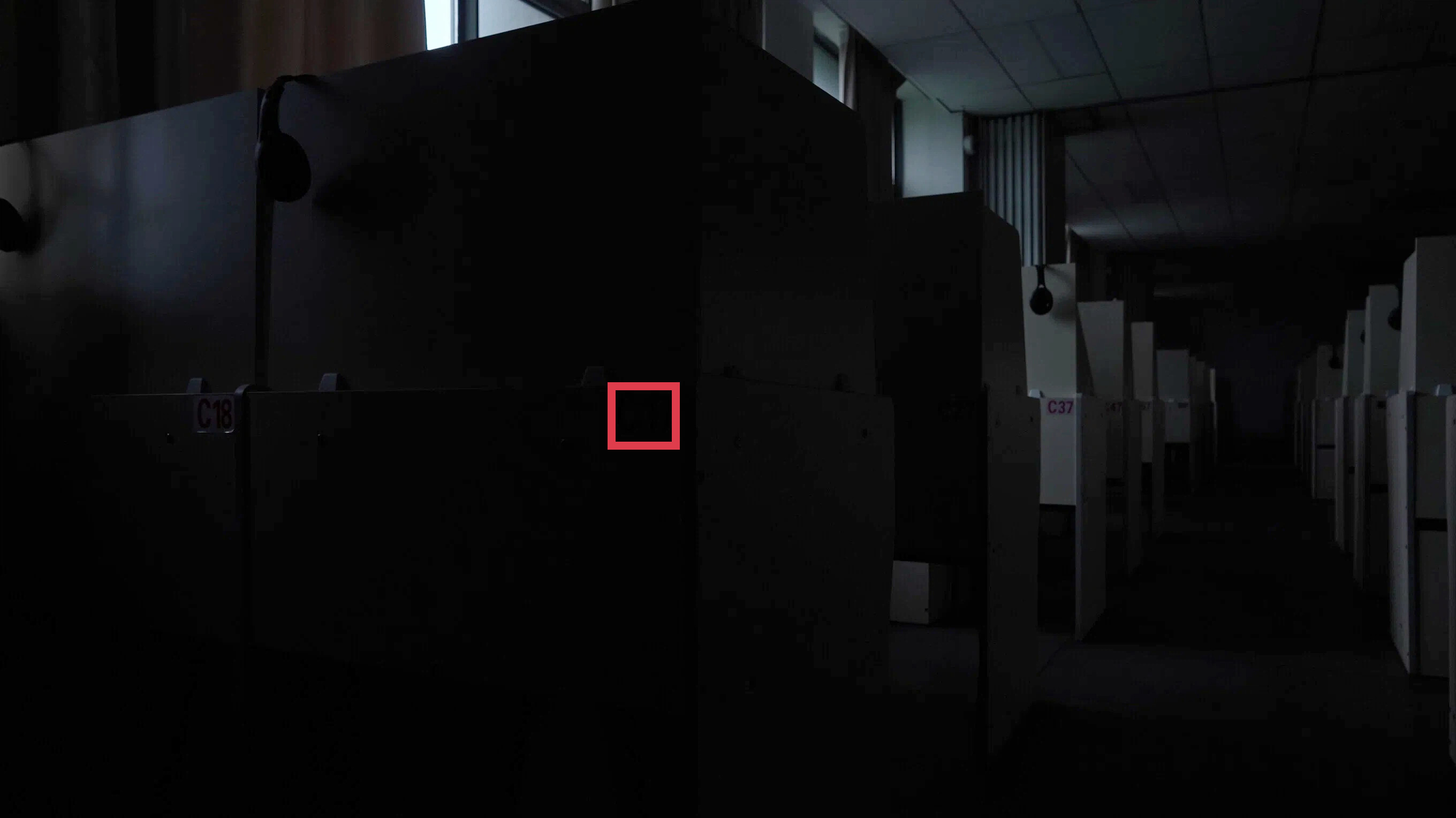}}}
            & \hspace{-4.0mm} \includegraphics[width=0.11\linewidth]{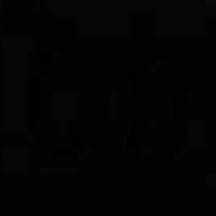}
            & \hspace{-4.0mm} \includegraphics[width=0.11\linewidth]{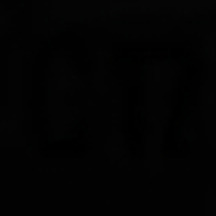} 
            & \hspace{-4.0mm} \includegraphics[width=0.11\linewidth]{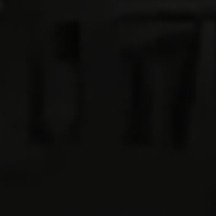} 
            & \hspace{-4.0mm} \includegraphics[width=0.11\linewidth]{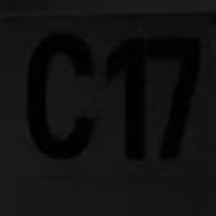} 
            & \hspace{-4.0mm} \includegraphics[width=0.11\linewidth]{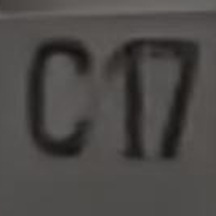} 
              \\
		\multicolumn{3}{c}{~}     
        & \hspace{-4.0mm} LQ patch 
        & \hspace{-4.0mm} PromptIR~\cite{PromptIR}
        & \hspace{-4.0mm} DA-CLIP~\cite{DA-CLIP}
        & \hspace{-4.0mm} DiffUIR~\cite{DiffUIR} 
        & \hspace{-4.0mm} AutoDIR~\cite{AutoDIR} \\  	
	\multicolumn{3}{c}{~} 
        & \hspace{-4.0mm} \includegraphics[width=0.11\linewidth]{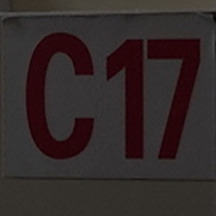}
        & \hspace{-4.0mm} \includegraphics[width=0.11\linewidth]{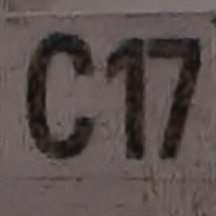} 
        & \hspace{-4.0mm} \includegraphics[width=0.11\linewidth]{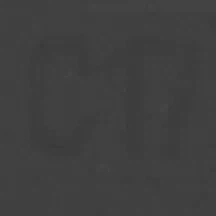} 
        & \hspace{-4.0mm} \includegraphics[width=0.11\linewidth]{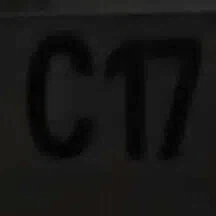} 
        & \hspace{-4.0mm} \includegraphics[width=0.11\linewidth]{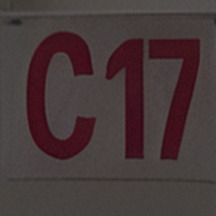} 

        \\
	\multicolumn{3}{c}{\hspace{-4.0mm} LQ image from FoundIR-TestData~\cite{foundir}} 
        & \hspace{-4.0mm} GT patch
        & \hspace{-4.0mm} InstructIR~\cite{InstructIR}
        & \hspace{-4.0mm} AgenticIR~\cite{AgenticIR}
        & \hspace{-4.0mm} FoundIR~\cite{foundir}
        & \hspace{-4.0mm} \textbf{FoundIR-v2} \\ 	

        \multicolumn{3}{c}{\multirow{5}*[48pt]{
            \hspace{-2.5mm} \includegraphics[width=0.42\linewidth,height=0.245\linewidth]{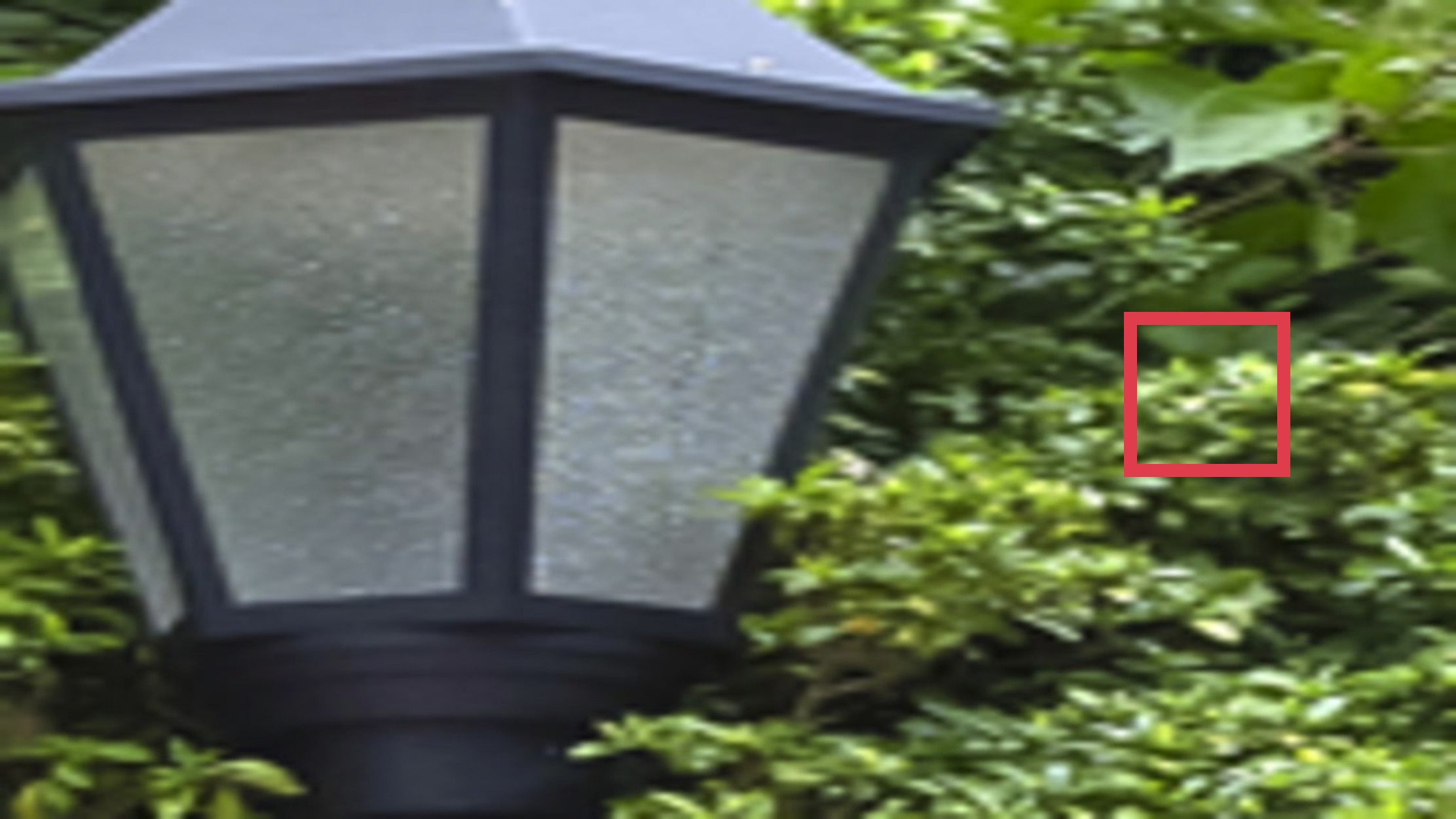}}}
            & \hspace{-4.0mm} \includegraphics[width=0.11\linewidth]{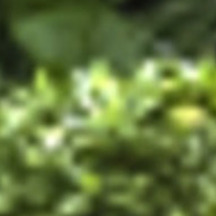}
            & \hspace{-4.0mm} \includegraphics[width=0.11\linewidth]{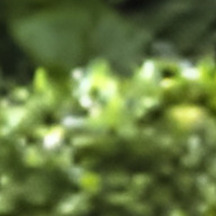} 
            & \hspace{-4.0mm} \includegraphics[width=0.11\linewidth]{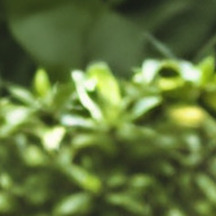} 
            & \hspace{-4.0mm} \includegraphics[width=0.11\linewidth]{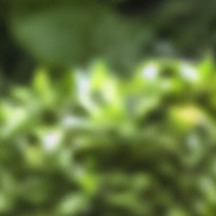} 
            & \hspace{-4.0mm} \includegraphics[width=0.11\linewidth]{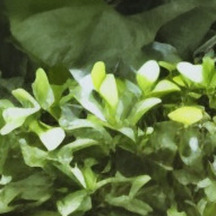} 
              \\
		\multicolumn{3}{c}{~}                                   & \hspace{-4.0mm} LQ patch
        & \hspace{-4.0mm} R-ESRGAN~\cite{Real-esrgan}
        & \hspace{-4.0mm} StabeSR~\cite{StabeSR} 
        & \hspace{-4.0mm} PASD~\cite{PASD}	
        & \hspace{-4.0mm} SeeSR~\cite{Seesr} \\
	\multicolumn{3}{c}{~} 
        & \hspace{-4.0mm} \includegraphics[width=0.11\linewidth]{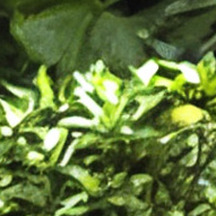} 
        & \hspace{-4.0mm} \includegraphics[width=0.11\linewidth]{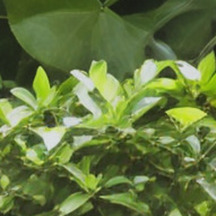} 
        & \hspace{-4.0mm} \includegraphics[width=0.11\linewidth]{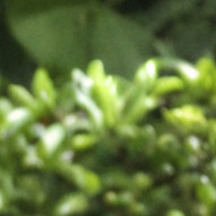} 
        & \hspace{-4.0mm} \includegraphics[width=0.11\linewidth]{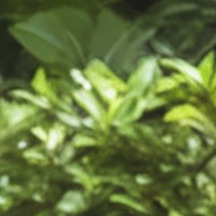} 
        & \hspace{-4.0mm} \includegraphics[width=0.11\linewidth]{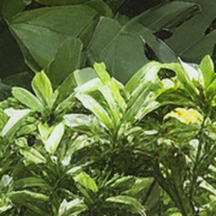}
        \\
	\multicolumn{3}{c}{\hspace{-4.0mm} LQ image from RealPhoto60~\cite{SUPIR}} 
        & \hspace{-4.0mm} DreamClear~\cite{DreamClear}
        & \hspace{-4.0mm} OSEDiff~\cite{OSEDiff} 
        & \hspace{-4.0mm} SUPIR~\cite{SUPIR}
        & \hspace{-4.0mm} FaithDiff~\cite{Faithdiff}
        & \hspace{-4.0mm} \textbf{FoundIR-v2}
        \\ 
		\end{tabular}
	\end{center}
	\vspace{-5mm}
	\caption{Visual comparison of image restoration results on the FoundIR-L+N and RealPhoto60 benchmarks. Zoom in for a better view.}
	\label{fig:visual}
	\vspace{-4mm}
\end{figure*}

{\flushleft\textbf{Evaluation on the public benchmarks}.}  
We first evaluate our FoundIR-v2 on public benchmark datasets across different restoration tasks and report the quantitative results in Table~\ref{tab:comparison}. 
We compare our method with state-of-the-art restoration methods, including all-in-one image restoration approaches (\emph{i.e.}, PromptIR~\cite{PromptIR}, TransWeather~\cite{Transweather}, DA-CLIP~\cite{DA-CLIP}, DiffUIR~\cite{DiffUIR}, AutoDIR~\cite{AutoDIR}, InstructIR~\cite{InstructIR}, X-Restormer~\cite{X-Restormer}, FoundIR~\cite{foundir}), and real-world super-resolution (SR) approaches (\emph{i.e.}, Real-ESRGAN~\cite{Real-esrgan}, BSRGAN~\cite{BSRGAN}, StabeSR~\cite{StabeSR}, PASD~\cite{PASD}, SeeSR~\cite{Seesr}, DreamClear~\cite{DreamClear}, OSEDiff~\cite{OSEDiff}, SUPIR~\cite{SUPIR}, FaithDiff~\cite{Faithdiff}).
Based on the statistics of five evaluation metrics on these test sets, our method achieves the best or second-best performance in more than 80\% of the quantitative comparisons.
Compared with existing all-in-one image restoration models, our FoundIR-v2 demonstrates more comprehensive restoration capabilities in complex and diverse scenarios.
Furthermore, compared with other real-world SR methods, our model also achieves significant improvements in terms of non-reference perceptual evaluations.
We further present several visual comparison results in Figure~\ref{fig:visual}. It can be clearly observed that the results of other methods still exhibit noticeable degradation, whereas our method performs better at simultaneously removing degradation and restoring details.

\begin{figure}[!t]
    \centering
    \begin{subfigure}{0.24\linewidth}
        \centering
        \includegraphics[width=\linewidth]{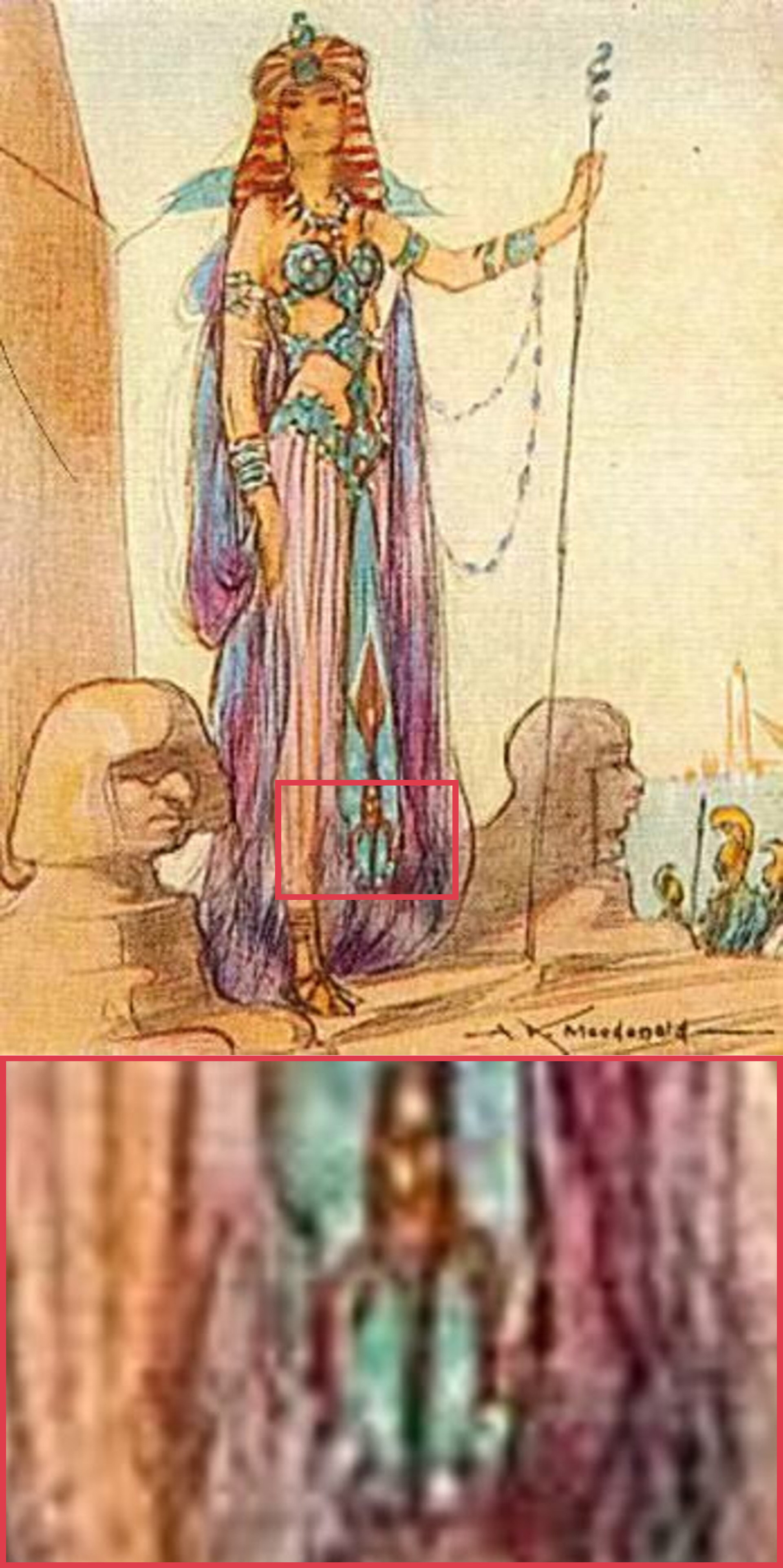}
        \caption{LQ}
        \label{fig:mural1}
    \end{subfigure}
    \begin{subfigure}{0.24\linewidth}
        \centering
        \includegraphics[width=\linewidth]{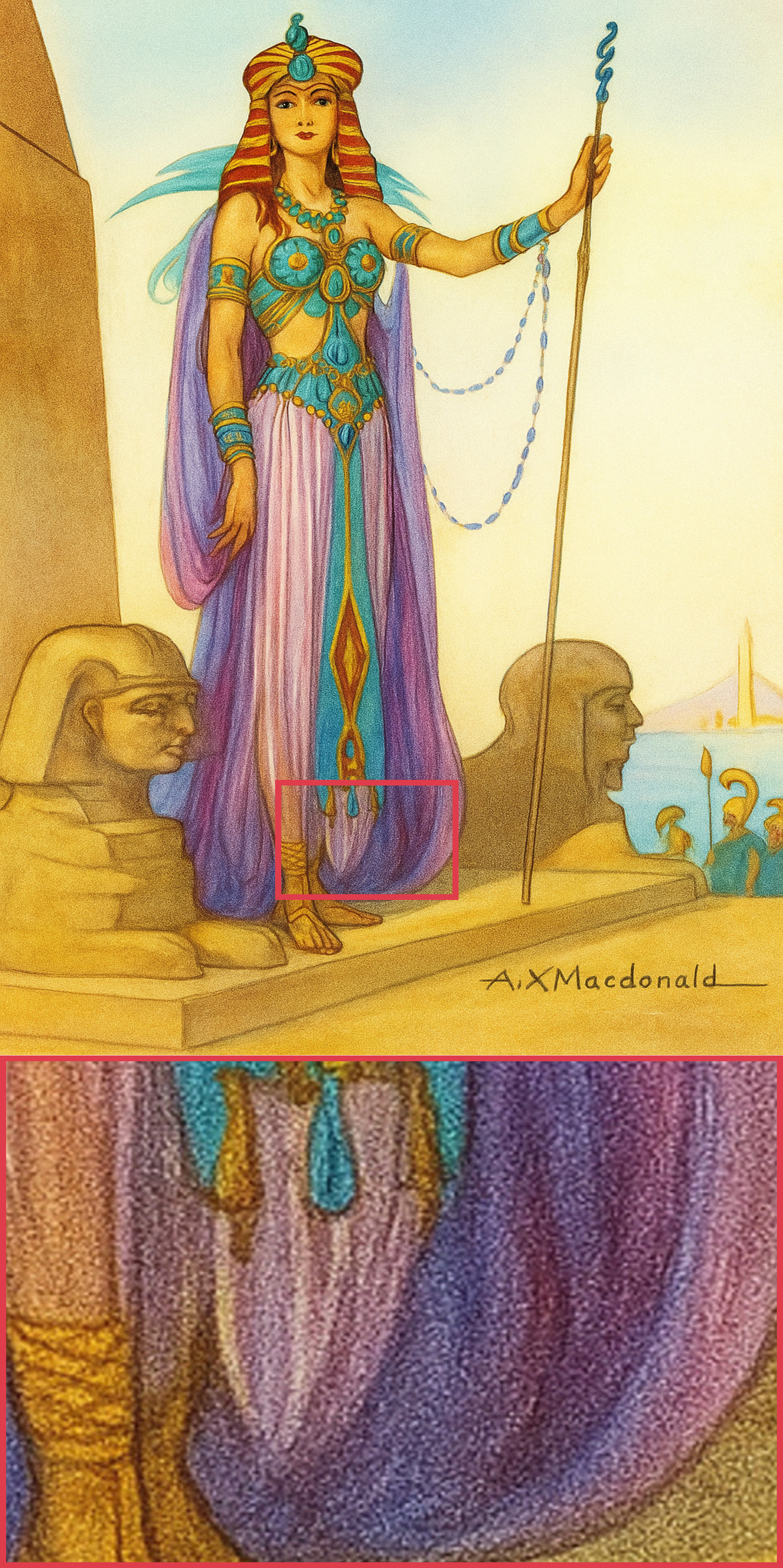}
        \caption{GPT-5}
        \label{fig:mural2}
    \end{subfigure}
    \begin{subfigure}{0.24\linewidth}
        \centering
        \includegraphics[width=\linewidth]{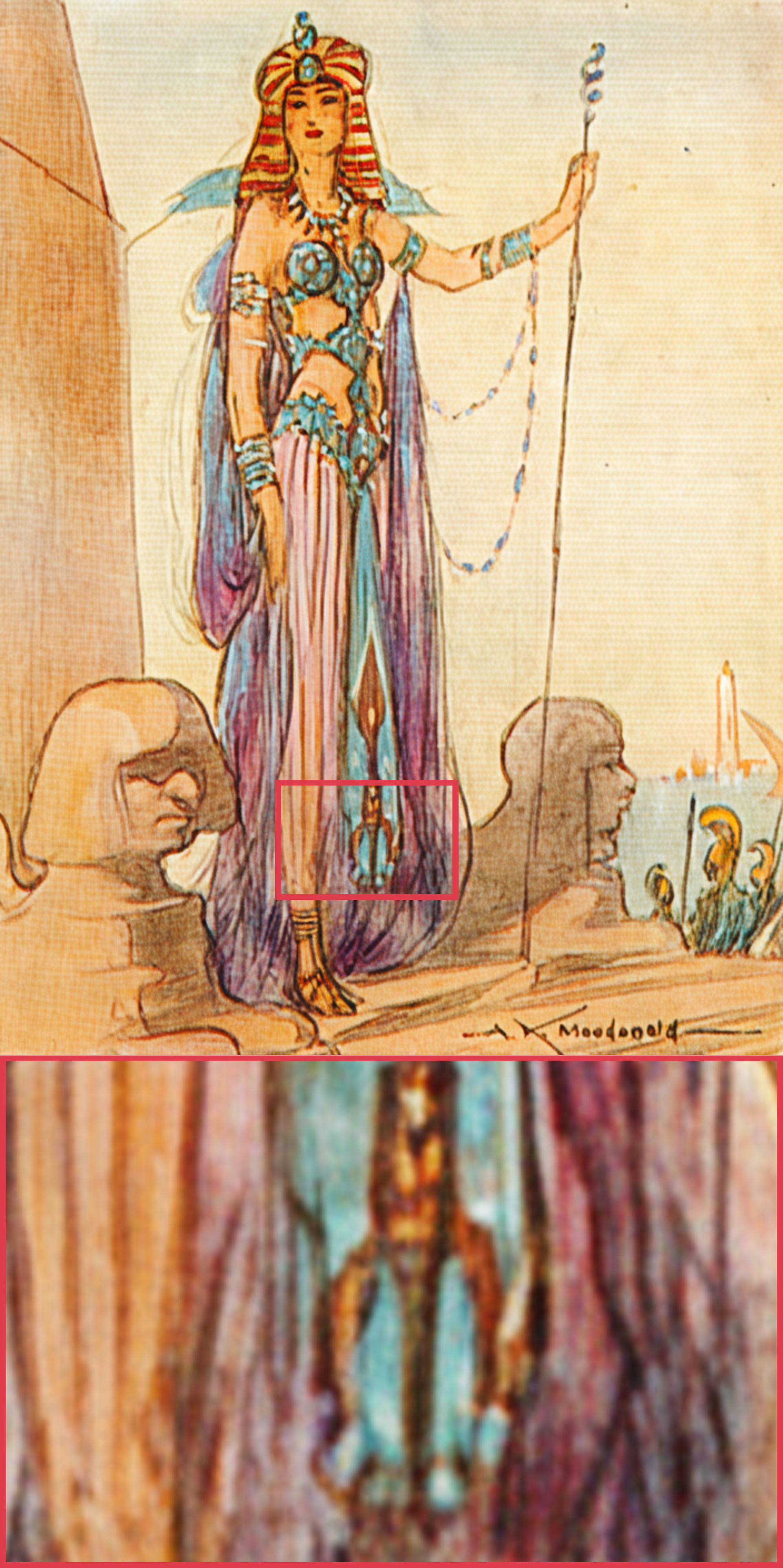}
        \caption{HYPIR~\cite{HYPIR}}
        \label{fig:mural3}
    \end{subfigure}
    \begin{subfigure}{0.24\linewidth}
        \centering
        \includegraphics[width=\linewidth]{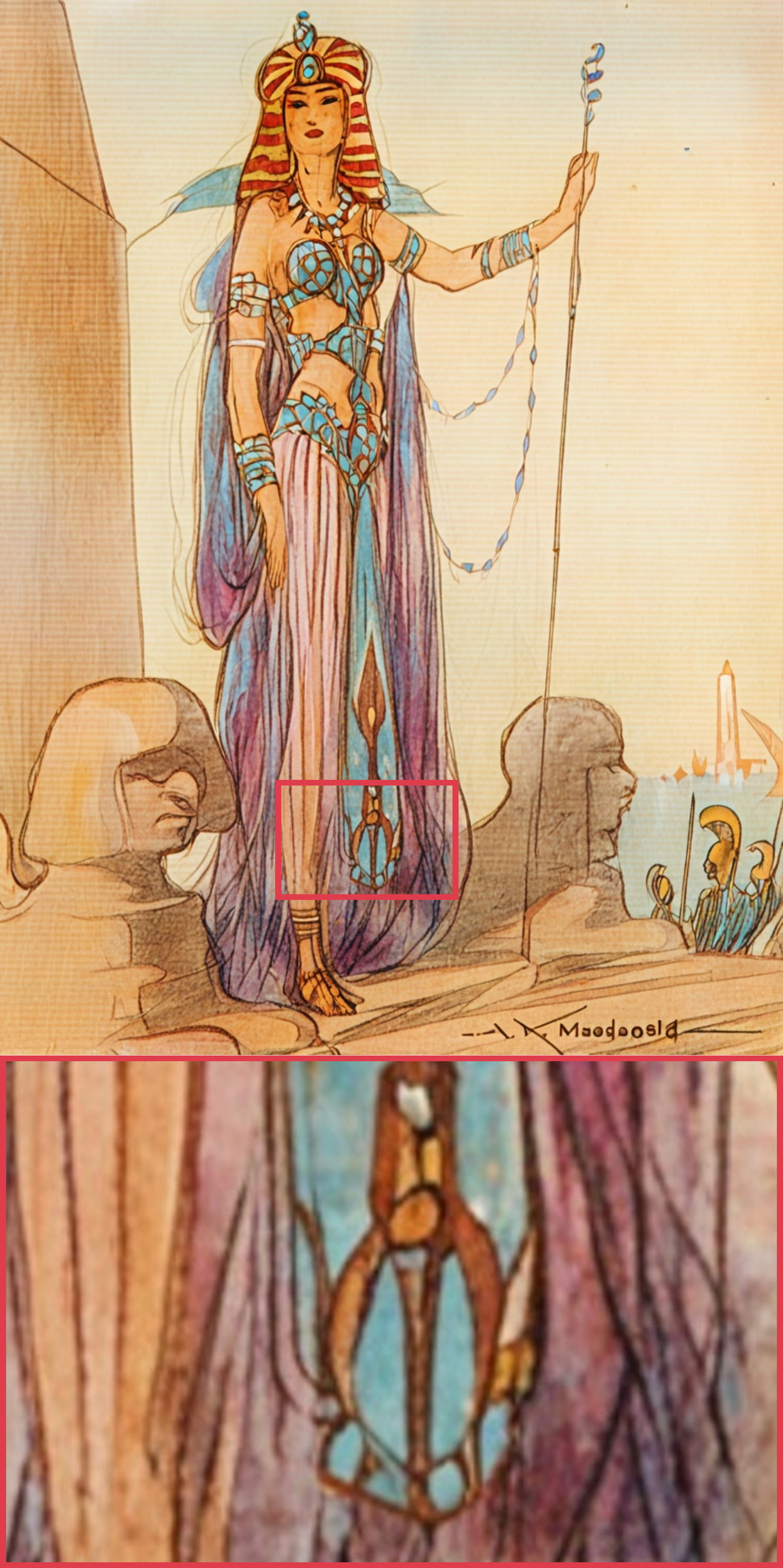}
        \caption{FoundIR-v2}
        \label{fig:mural4}
    \end{subfigure}
    \vspace{-2mm}
    \caption{Visual comparison of mural restoration results.}
    \label{fig:mural}
    \vspace{-2mm}
\end{figure}

\vspace{-2mm}

{\flushleft\textbf{Comparison with image restoration agent}.}  
We also compare our approach with the recent image restoration agent, \emph{e.g.}, AgenticIR~\cite{AgenticIR}.
Compared with AgenticIR that intelligently schedules multiple specialized restoration models, our FoundIR-v2 achieves better reconstruction quality, especially in coupled degradation tasks (Table~\ref{tab:comparison} and Figure~\ref{fig:visual}).

\vspace{-2mm}

{\flushleft\textbf{Comparison with commercial methods}.}  
We further compare our method with two popular commercial large-scale models, \emph{i.e.}, GPT-5 and HYPIR~\cite{HYPIR}.
To investigate the potential of these models when applied to challenging tasks involving unknown and complex degradations, we conduct experiments on the mural restoration task, where the degradations are often highly uncertain and heterogeneous~\cite{shao2023building}.
As presented in Figure~\ref{fig:mural}, GPT-5 struggles to maintain pixel-level structural fidelity, and the generated content exhibits noticeable deviations from the original mural style.
Furthermore, the recovery results of HYPIR still contain some undesirable noise. In contrast, our method produces faithful and visually pleasing results, demonstrating the generalization ability of FoundIR-v2 on unknown challenging tasks. 

\subsection{Ablation analysis and discussion}
In this section, we conduct extensive ablation studies. Due to considerations of training costs, we select four representative sub-tasks (\emph{i.e.}, deblurring, dehazing, low-light enhancement, and SR) for all-in-one performance analysis.

\vspace{-2mm}

{\flushleft\textbf{Effect of data mixture proportion}.}  
To analyze the effect of different data mixing proportions in the training batch, we train models with varying proportions under the same settings for fair comparisons. 
Figure~\ref{proportion} demonstrates that training with equal data proportions does not yield optimal performance, as different tasks exhibit varying levels of learning difficulty.
These results provide valuable insights into the training of image restoration foundation models, highlighting the importance of optimizing data mixture.

\vspace{-2mm}

{\flushleft\textbf{Effect of different training strategies on the same model}.} 
We compare our data equilibrium scheduling (DES) strategy with other all-in-one training approaches, including (1) mixed training~\cite{PromptIR}, (2) multi-task sequential learning~\cite{kong2024towards}, and (3) multi-task incremental learning~\cite{foundir}.
Table~\ref{tab:training} reports the average quantitative results of different variants across the four sub-tasks.
Note that we retrain all comparison models for a fair comparison.
We can observe that for FoundIR-v2 as a fixed model, our DES strategy achieves the best performance, outperforming other training strategies.

Furthermore, we apply our proposed DES strategy to existing all-in-one restoration models, \emph{i.e.}, PromptIR~\cite{PromptIR} and FoundIR~\cite{foundir}. 
Compared with their respective original fixed-ratio training schemes, our method brings performance gains by dynamically optimizing data mixture.

\vspace{-2mm}

\begin{table}[!t]
\footnotesize
\centering
\caption{Ablation analysis on different all-in-one image restoration models under different training strategies. ``Our*'' indicates that our FoundIR-v2 is trained without the SR task for a fair comparison, since PromptIR and FoundIR do not include the SR task.}
\vspace{-2mm}
\resizebox{0.5\textwidth}{!}{%
    \begin{tabular}{l|cccc|cc}
        \bottomrule
        \multirow{2}{*}{\makecell{Base Models}} & \multicolumn{4}{c|}{Training Strategy}        & \multicolumn{2}{c}{Metrics} \\ \cline{2-7} 
                                      & Mixing & Sequence & Incremental & DES (Ours) & PSNR~$\uparrow$         & SSIM~$\uparrow$         \\ \hline
        \multirow{2}{*}{PromptIR~\cite{PromptIR}}   & \CheckmarkBold      & \XSolidBrush         & \XSolidBrush            & \XSolidBrush           & 15.89        & 0.6314       \\
                                    & \XSolidBrush        & \XSolidBrush         & \XSolidBrush            & \CheckmarkBold         & 16.92        & 0.6888       \\ \hline
        \multirow{2}{*}{FoundIR~\cite{foundir}}    & \XSolidBrush        & \XSolidBrush         & \CheckmarkBold          & \XSolidBrush           & 16.77        & 0.6880       \\
                                    & \XSolidBrush        & \XSolidBrush         & \XSolidBrush            & \CheckmarkBold         & 17.95        & 0.7312       \\ \hline
        \multirow{1}{*}{Ours*}      & \XSolidBrush        & \XSolidBrush         & \XSolidBrush            & \CheckmarkBold         & 20.09        & 0.7447       \\ \hline
        \multirow{4}{*}{Ours}       & \CheckmarkBold      & \XSolidBrush         & \XSolidBrush            & \XSolidBrush           & 18.91        & 0.6759       \\
                                    & \XSolidBrush        & \CheckmarkBold       & \XSolidBrush            & \XSolidBrush           & 18.69        & 0.6476       \\
                                    & \XSolidBrush        & \XSolidBrush         & \CheckmarkBold          & \XSolidBrush           & 19.93        & 0.6725       \\
                                    & \XSolidBrush        & \XSolidBrush         & \XSolidBrush            & \CheckmarkBold         & 20.41        & 0.6977       \\ \bottomrule
    \end{tabular}
}
\label{tab:training}
\end{table}

{\flushleft\textbf{Effect of different models on the same training strategy}.} 
To disentangle the interplay between the effectiveness of training strategies and that of models, we further analyze the effect of different models under the same training strategy.
Under the same DES strategy, our FoundIR-v2 achieves a significant performance improvement over FoundIR.
This confirms that, compared with FoundIR’s diffusion model learned in image space, our FoundIR-v2 leverages diffusion priors in latent space to better facilitate image restoration.

\begin{figure*}[!t]
    \centering
    \begin{minipage}{\linewidth}
        \centering
        \begin{subfigure}{0.16\linewidth}
            \includegraphics[width=\linewidth]{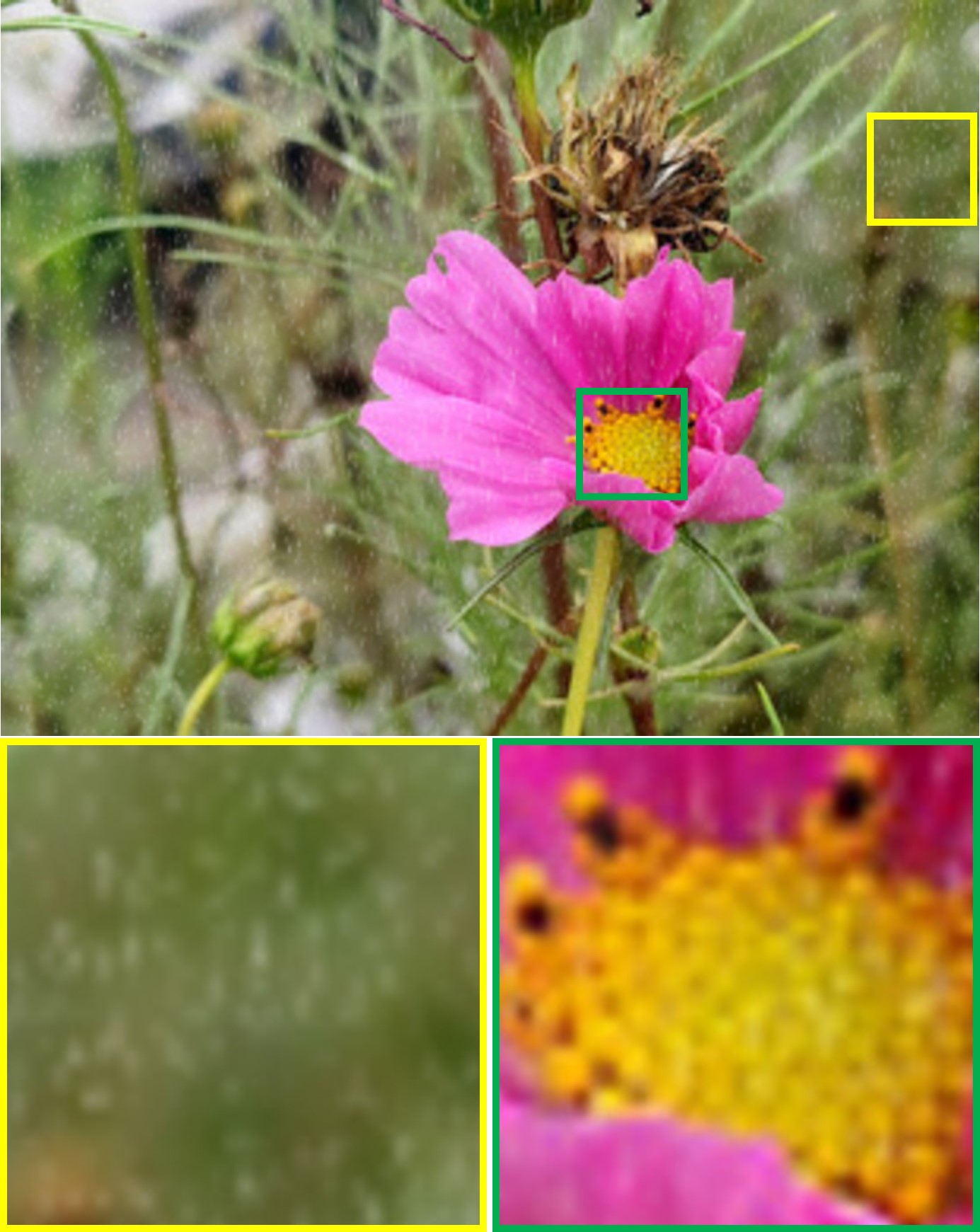}
            \caption{LQ}
        \end{subfigure}
        \begin{subfigure}{0.16\linewidth}
            \includegraphics[width=\linewidth]{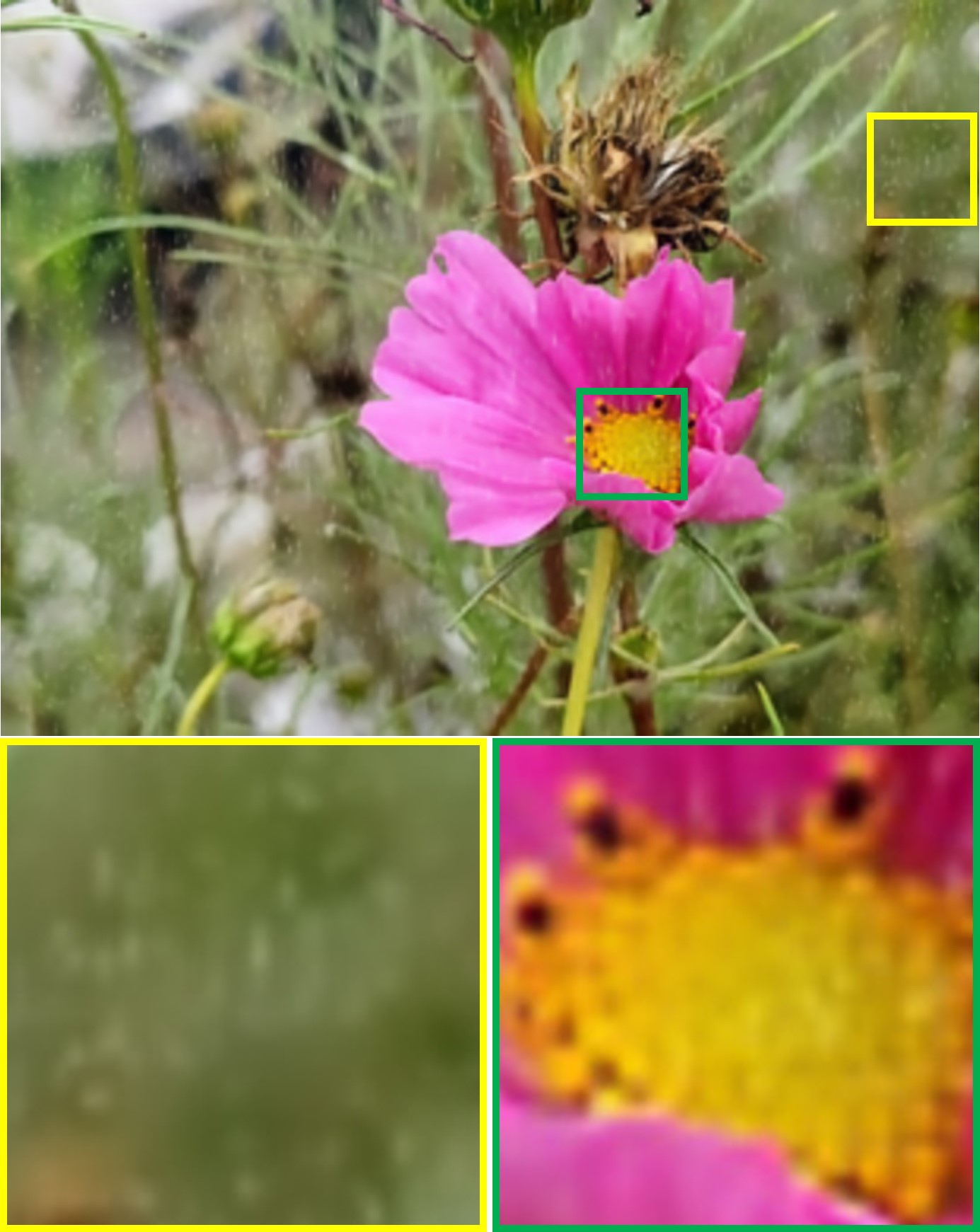}
            \caption{FoundIR}
        \end{subfigure}
        \begin{subfigure}{0.16\linewidth}
            \includegraphics[width=\linewidth]{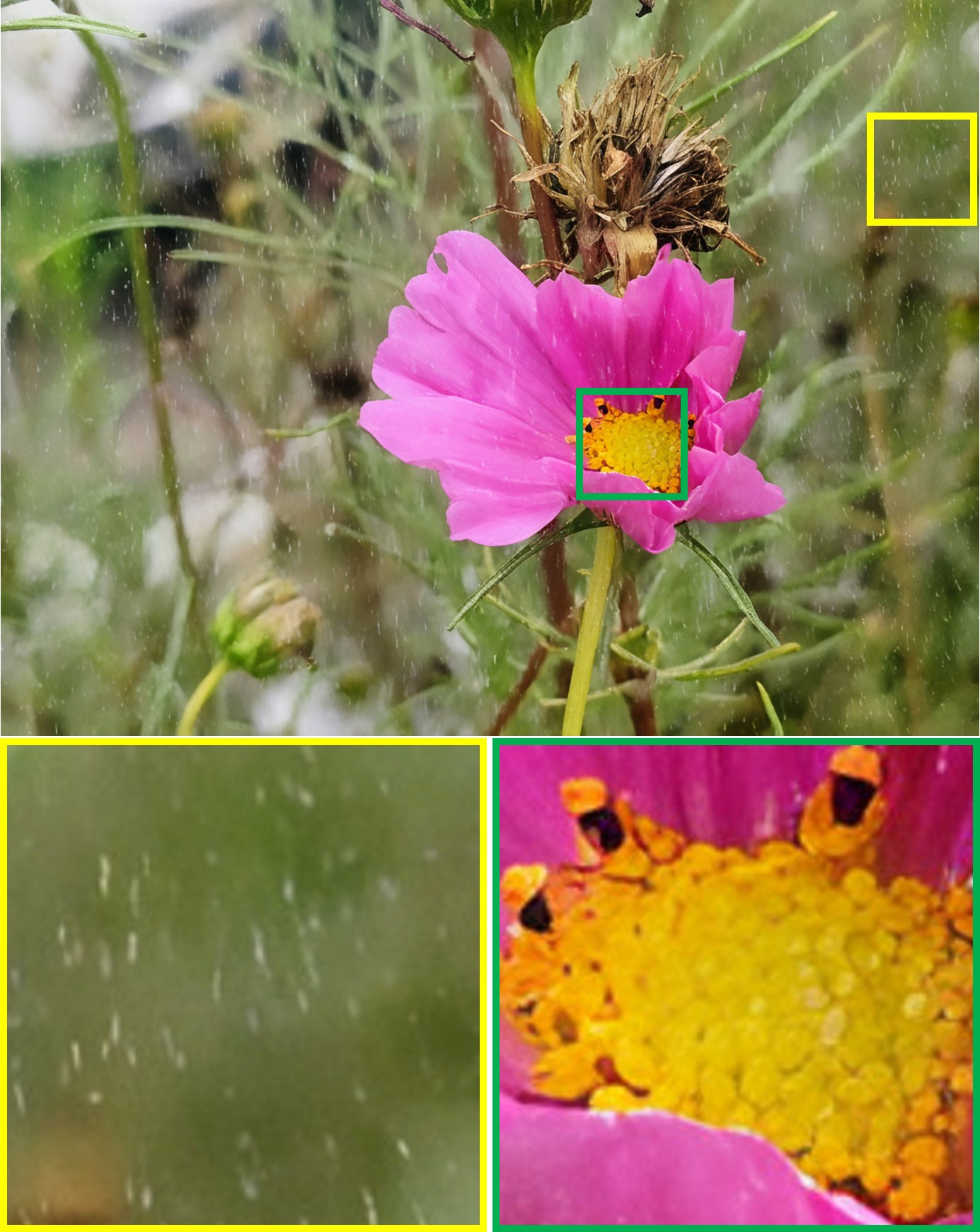}
            \caption{FoundIR+SUPIR}
        \end{subfigure}
        \begin{subfigure}{0.16\linewidth}
            \includegraphics[width=\linewidth]{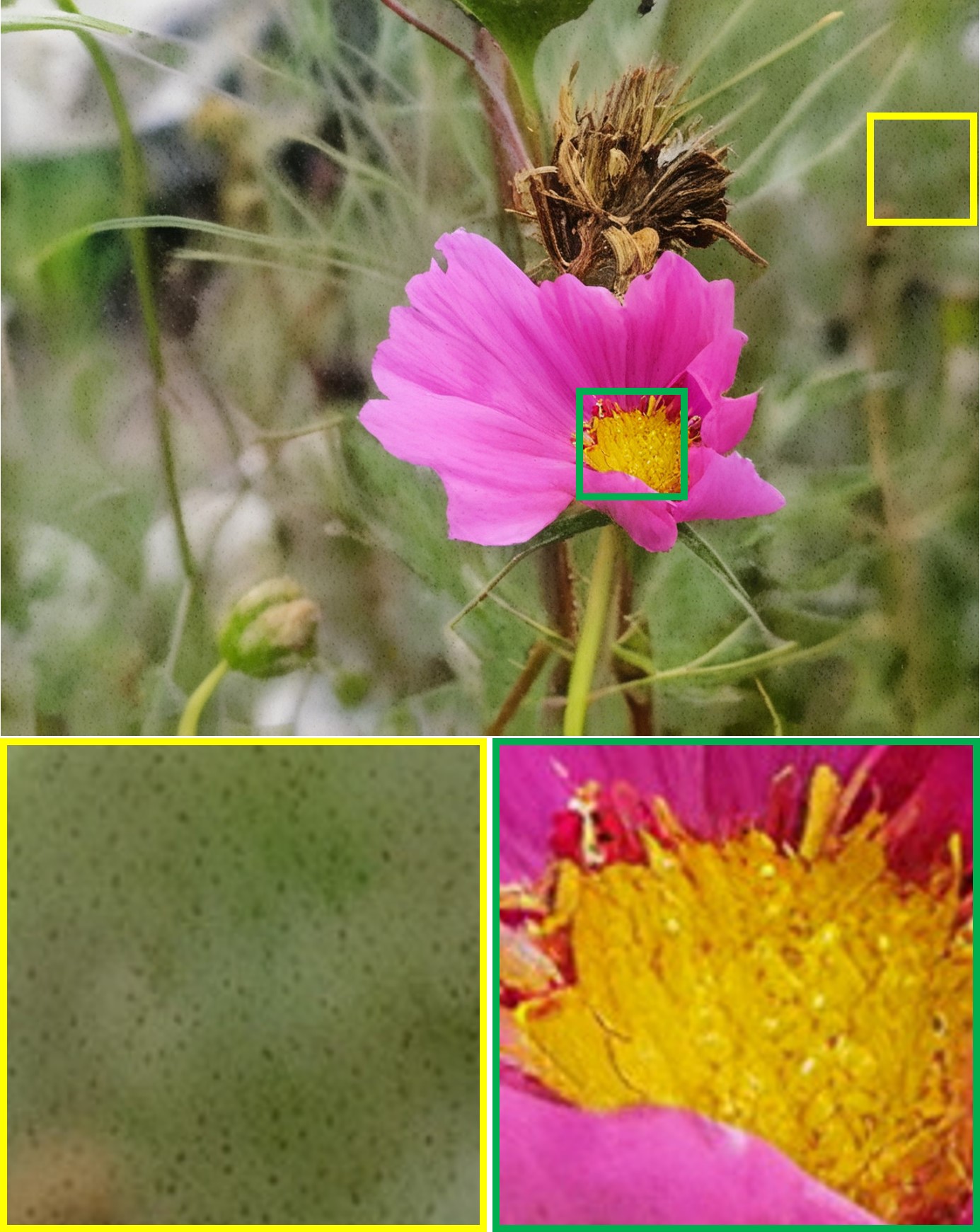}
            \caption{SUPIR+FoundIR}
        \end{subfigure}
        \begin{subfigure}{0.16\linewidth}
            \includegraphics[width=\linewidth]{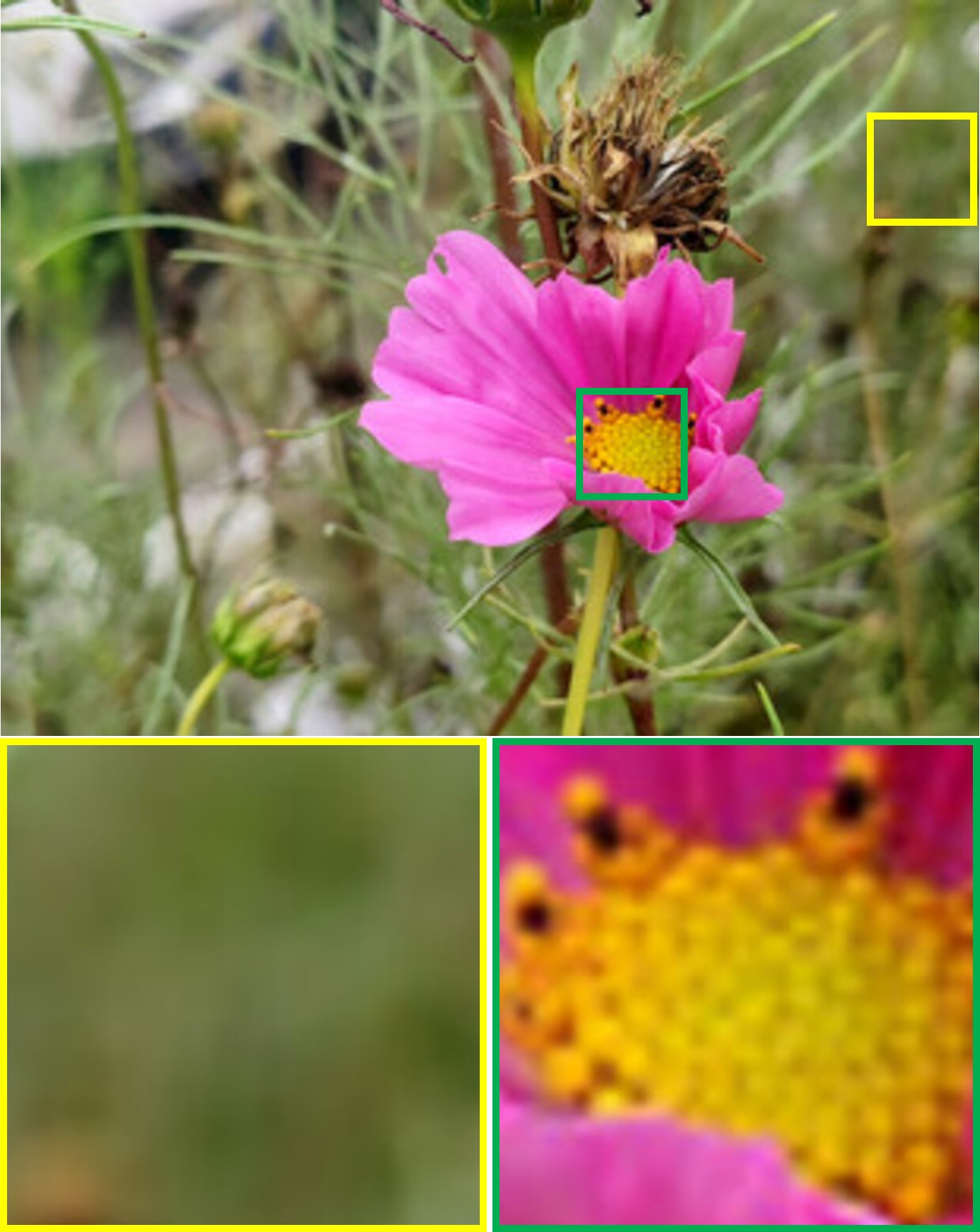}
            \caption{Ours (w/o SR)}
        \end{subfigure}
        \begin{subfigure}{0.16\linewidth}
            \includegraphics[width=\linewidth]{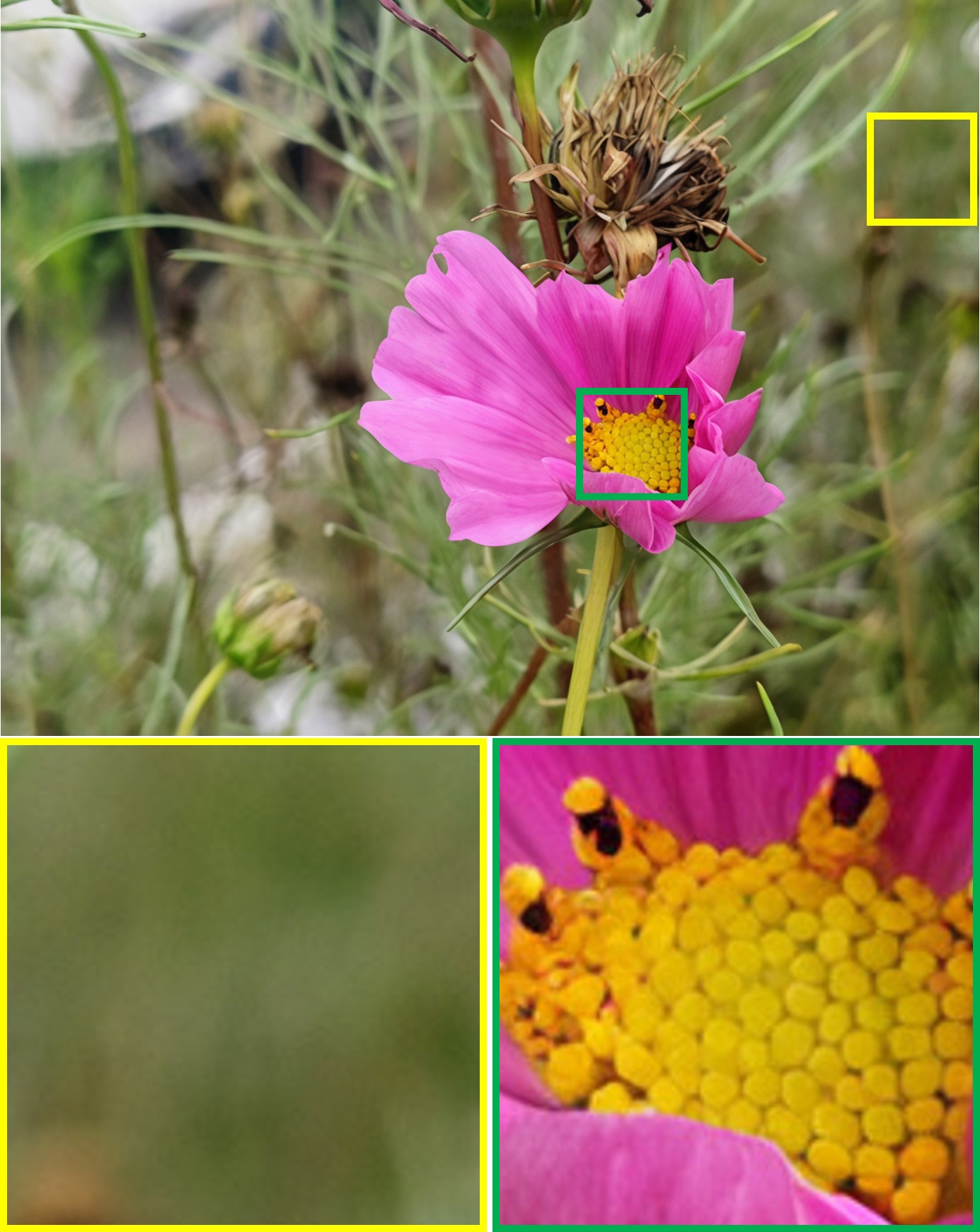}
            \caption{Ours}
        \end{subfigure}
    \end{minipage}
    \vspace{-3mm}
    \caption{Visual comparison results for low-resolution image restoration. Compared to the cascaded use of FoundIR~\cite{foundir} (all-in-one model w/o SR task) and SUPIR~\cite{SUPIR} (SR), our FoundIR-v2 integrates both degradation removal and detail generation capabilities simultaneously.}
    \label{fig:rainsr}
    \vspace{-4mm}
\end{figure*}

\vspace{-2mm}

\begin{figure}[!t]
    \centering
    \begin{subfigure}{0.48\linewidth}
        \centering
        \includegraphics[width=\linewidth]{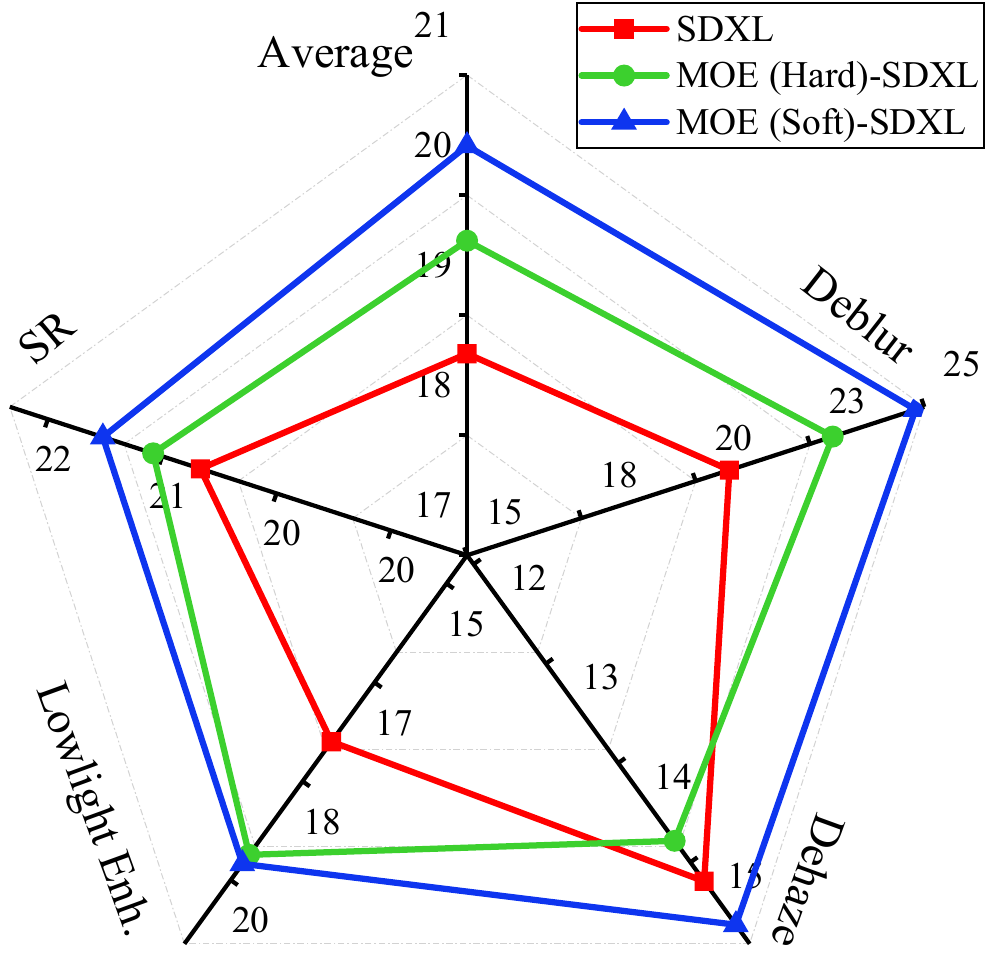}
        \caption{MoE-driven scheduler}
        \label{fig:MOE}
    \end{subfigure}
    \begin{subfigure}{0.5\linewidth}
        \centering
        \includegraphics[width=\linewidth]{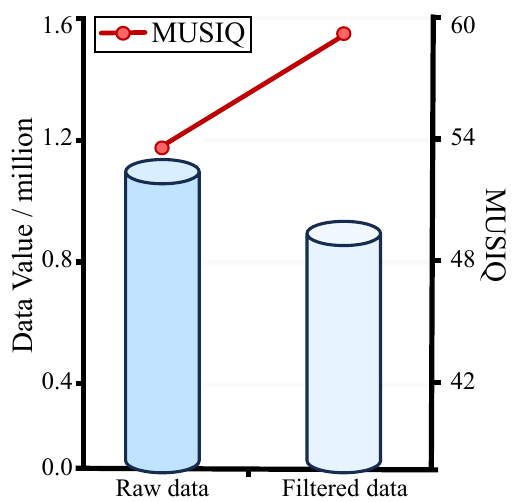}
        \caption{GT data cleaning}
        \label{fig:GT}
    \end{subfigure}
    \vspace{-2mm}
    \caption{Statistical ablation analysis for the MoE-driven diffusion scheduler and GT data cleaning.}
    \label{fig:MOEGT}
\end{figure}

{\flushleft\textbf{Effectiveness of MoE-driven diffusion scheduler}.}  
To validate the effectiveness of our proposed MoE-driven diffusion scheduler, we compare our architecture with other variants, including (i) without scheduler (\emph{i.e.}, only SDXL), and (ii) hard MOE-based scheduler~\cite{MoCE} (\emph{i.e.}, only top-k expert).
We present the radar chart in Figure~\ref{fig:MOEGT}(a) to illustrate the restoration capabilities of different variants across multiple tasks.
It can be observed that our designed scheduler with the soft MoE consistently obtains higher performance, as the soft gating mechanism enables adaptive collaboration among experts for better all-in-one image restoration.

\vspace{-2.5mm}

{\flushleft\textbf{Effectiveness of GT data cleaning}.}  
To demonstrate the effectiveness of GT data cleaning, we compare our approach with directly combining all datasets for training foundation model.
As evidenced by the results in Figure~\ref{fig:MOEGT}(b), filtering out low-quality data from the original all-in-one training dataset can further improve the restoration performance.
This also highlights that the quality of GTs is equally important for image restoration foundation model.

\vspace{-2.5mm}

{\flushleft\textbf{Discussion with other potential foundation models}.}
We note that different universal restoration models cover different ranges of tasks.
For example, for low-resolution rainy image, FoundIR~\cite{foundir} can perform deraining but lacks SR ability, whereas SUPIR~\cite{SUPIR} can perform SR but does not handle deraining.
Compared with these all-in-one or universal restoration methods, our approach consolidates more restoration capabilities into a unified foundation model.
We present an example in Figure~\ref{fig:rainsr}.
In contrast to cascadely applying FoundIR~\cite{foundir} and SUPIR~\cite{SUPIR}, our model produces more visually appealing results by performing rain removal and SR simultaneously.
This further demonstrates that our method better leverages the synergy among different tasks.

\vspace{-2mm}

\begin{figure}[!t]
    \centering
    \begin{minipage}{\linewidth}
        \centering
        \begin{subfigure}{0.325\linewidth}
            \includegraphics[width=\linewidth]{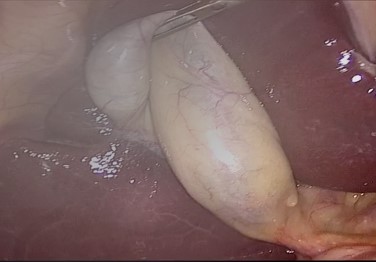}
        \end{subfigure}
        \begin{subfigure}{0.325\linewidth}
            \includegraphics[width=\linewidth]{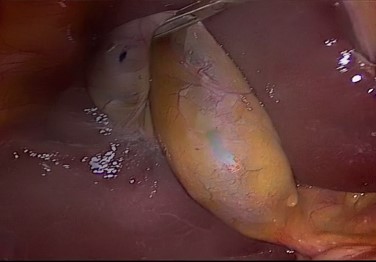}
        \end{subfigure}
        \begin{subfigure}{0.325\linewidth}
            \includegraphics[width=\linewidth]{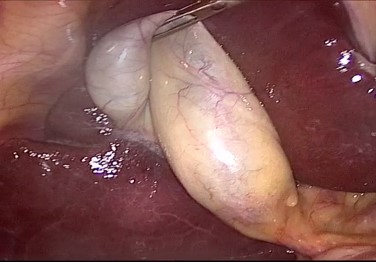}
        \end{subfigure}
    \end{minipage}
    \begin{minipage}{\linewidth}
        \centering
        \begin{subfigure}{0.325\linewidth}
            \includegraphics[width=\linewidth]{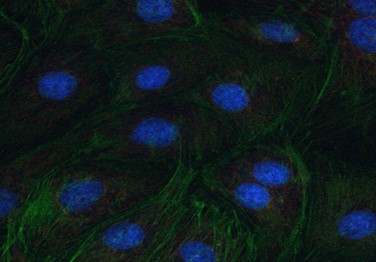}
            \caption{LQ}
        \end{subfigure}
        \begin{subfigure}{0.325\linewidth}
            \includegraphics[width=\linewidth]{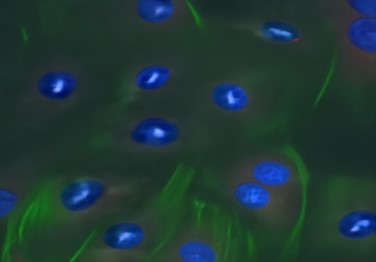}
            \caption{FoundIR~\cite{foundir}}
        \end{subfigure}
        \begin{subfigure}{0.325\linewidth}
            \includegraphics[width=\linewidth]{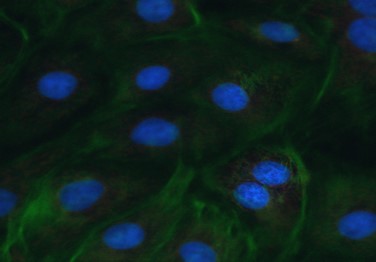}
            \caption{FoundIR-v2 (Ours)}
        \end{subfigure}
    \end{minipage}
    \vspace{-3mm}
    \caption{Visual comparison of fine-tuning the image restoration foundation models on downstream restoration tasks (laparoscopic surgery and biological microscopy image restoration).}
    \label{fig:applicaton}
\end{figure}

{\flushleft\textbf{Applications on downstream tasks}.}  
\label{sec4.3}
In fact, a well-trained image restoration foundation model serves as a versatile basis that can be fine-tuned for various downstream restoration tasks related to its pre-training objective.
To demonstrate the applicability of image restoration foundation models to downstream tasks, we extend FoundIR~\cite{foundir} and FoundIR-v2 to laparoscopic surgery and biological microscopy image restoration.
Since the training dataset in these tasks are typically limited and protected by privacy constraints, we fine-tune the pre-trained foundation models using only using a small amount of available data.
The qualitative comparison results are provided in Figure~\ref{fig:applicaton}.
Compared with FoundIR, our FoundIR-v2 achieves better reconstruction results on these downstream tasks, highlighting the significance of developing effective image restoration foundation models.

\section{Conclusion}
We have presented an effective image restoration foundation model, named FoundIR-v2.
We formulate a data equilibrium scheduling paradigm that dynamically optimizes all-in-one training data distributions to improve multi-task image restoration performance.
Furthermore, we integrate an MoE-driven scheduler into generative pre-training to better unleash task-adaptive diffusion priors for universal image restoration.
Extensive experiments and ablation analysis reveal the value of optimizing data mixture, and demonstrate the effectiveness of the proposed method. 

{
    \small
    \bibliographystyle{ieeenat_fullname}
    \bibliography{main}
}

\end{document}